\documentclass{article}

\PassOptionsToPackage{numbers, compress}{natbib}

     \usepackage[preprint]{neurips_2019}

\usepackage[utf8]{inputenc} 
\usepackage[T1]{fontenc}    
\usepackage{hyperref}       
\usepackage{url}            
\usepackage{booktabs}       
\usepackage{amsfonts}       
\usepackage{nicefrac}       
\usepackage{microtype}      

\usepackage{math_defs}
\usepackage{amsmath}
\usepackage{amssymb}
\usepackage{color}
\usepackage{enumitem}
\usepackage{mathrsfs}
\usepackage{graphicx}
\usepackage{subcaption}

\title{Adaptive Deep Kernel Learning}

\author{%
  Prudencio Tossou\thanks{} \\
  InVivo AI, Université Laval\\
  \texttt{prudencio@invivoai.com} \\
  \And
  Basile Dura\\
  InVivo AI, Mila\\
  \texttt{basile@invivoai.ca} \\
   \And
   Mario Marchand \\
   Université Laval \\
   \texttt{mario.marchand@ift.ulaval.ca} \\
   \AND
   François Laviolette \\
   Université Laval \\
   \texttt{francois.laviolette@ift.ulaval.ca} \\
   \And
   Alexandre Lacoste \\
   Element AI \\
   \texttt{allac@elementai.com} \\
}

\begin{document}

\maketitle

\begin{abstract}
    Deep kernel learning provides an elegant and principled framework for combining the structural properties of deep learning algorithms with the flexibility of kernel methods.
    By means of a deep neural network, we learn a parametrized kernel operator that can be combined with a differentiable kernel algorithm during inference.
    While previous work within this framework has focused on learning a single kernel for large datasets, we learn a kernel family for a variety of few-shot regression tasks.
    Compared to single deep kernel learning, our algorithm enables identification of the appropriate kernel for each task during inference.
    As such, it is well adapted for complex task distributions in a few-shot learning setting, which we demonstrate by comparing against existing state-of-the-art algorithms using real world, few-shot regression tasks related to the field of drug discovery.
\end{abstract}

\section{Introduction}
Recent studies exploring few-shot learning have led to the development of new algorithms that can learn efficiently from a small number of samples and generalize beyond the training data \citep{yaqing2019review, chen2019closer}.
Most of these algorithms have adopted the meta-learning paradigm \citep{thrun1998learning, vilalta2002perspective}, where prior knowledge is learned across a large collection of diverse tasks and then transferred to new tasks for efficient learning with limited amount of data.
Despite many few-shot learning algorithms reporting improved performance over previous iterations, considerable progress is needed before these approaches can be adopted at scale and in practical settings.
Notably, much of the work in this field has focused on classification and reinforcement learning \cite{yaqing2019review}, leaving the problem of few-shot regression largely unaddressed \cite{li2017meta, kim2018bayesian, Loo2019regression}.
In addition, to the best of our knowledge, no real world few-shot regression (FSR) benchmarks have been established in the literature.
However, regression methods would be widely applicable to important problems encountered in multiple fields plagued by small sample sizes.
As an example, molecules in drug discovery often have their properties quantified in a dose-dependant manner, a process that is expensive and time consuming, limiting the amount of examples available in each discovery program.

Existing meta-learning-based few-shot algorithms differ by the nature of the meta-knowledge captured and the amount of adaptation performed at test-time for new tasks or data sets.
First, metric learning methods \citep{koch2015siamese, vinyals2016matching, snell2017prototypical, garcia2017few, bertinetto2018meta} accumulate meta-knowledge in high capacity covariance or distance functions that are combined with simple base learners to make predictions.
One caveat of these methods is that they do not adapt their covariance functions at test-time. Hence, few of the currently-used base learners have enough capacity to adapt appropriately \citep{bertinetto2018meta, triantafillou2019meta} to new tasks.
Second, initialization and optimization based methods \cite{finn2017model, kim2018bayesian, ravi2016optimization} that learn the initialization point for gradient descent algorithms
allow for more adaptation on new tasks, but remain time consuming and memory inefficient.
To ensure optimal performance on FSR problems, we propose combining the strengths of both approaches.

In this study, we frame FSR as a deep kernel learning (DKL) problem as opposed to one of metric learning, allowing us to derive new algorithms.
DKL methods combine the non-parametric flexibility of kernel methods with the structural properties of deep neural networks, which yields a more powerful method for learning input covariance functions and achieve greater adaptation capacity at test-time.
We offer further improvements over the general DKL algorithm for FSR by learning a family of covariance functions and selecting the most appropriate one at test-time, resulting in greater adaptability.

\textbf{Our Contributions:} First, we frame few-shot regression as a deep kernel learning problem and explain why it is more expressive than classical metric learning methods and how this is achieved.
Next, we derive two DKL algorithms by combining set embedding techniques \cite{zaheer2017deep} and kernel methods \cite{scholkopf2001learning, williams1996gaussian} to learn a family of kernels, allowing more test-time adaptation while being sample efficient.
We then propose two new real-world datasets drawn from the drug discovery domain for FSR.
Performance on these datasets as well as synthetic data shows that our model allows better few-shot training and generalization than previously proposed methods.

\section{Preliminaries}\label{sec:preliminaries}
In this section, we describe in depth the DKL framework (introduced by \citet{wilson2016deep}) and show that it can be adapted to learn a covariance or kernel function for few-shot learning tasks.

\paragraph{DKL Framework:} Let $D^t_{trn} = \LC (\xb_i, y_i) \RC_{i=1}^m \subset \Xcal \times \Reals$, a training dataset available for learning the regression task $t$  where $\Xcal$ is the input space and $\Reals$ is the output space.
A DKL algorithm aims to obtain a non-linear embedding of inputs in the embedding space $\Hcal$, using a deep neural network $ \phib_{\thetab}: \Xcal \to \Hcal$ of parameters $\thetab$.
Then, it finds the minimal norm regressor $ h^t_* $ in the reproducing kernel Hilbert space (RKHS) $\Rcal$
on $\Hcal$, that fits the training data, i.e.:
\begin{equation}\label{eq:minimal_norm_regression}
    h^t_* := \argmin{h \in \Rcal} \lambda \norm{h}_{\Rcal} + \ell (h, \Dcal^t_{trn})
\end{equation}
where $\ell$ is a non-negative loss function that measures the loss of a regressor $h$ and $\lambda$ weighs the importance of the norm minimization against the training loss.
Following the representer theorem \citep{scholkopf2001learning, steinwart2008support}, $h^t_*$ can be written as a finite linear combination of kernel evaluations on training inputs, i.e.:
\begin{equation}\label{eq:ht_dual}
h^t_*(\xb) = \sum_{(\xb_i, y_i) \in D^t_{trn}}{\alpha^t_i  k_{\rhob}(\phib_{\thetab}(\xb), \phib_{\thetab}(\xb_i))},
\end{equation}
where $\alb^t = (\al^t_1, \cdots, \al^t_m)$ are the combination weights and $k_{\rhob} \colon \Hcal \times \Hcal \to \Reals_+$ is a reproducing kernel of $\Rcal$ with hyperparameters $\rhob$.
Candidates include the radial basis, polynomial, and linear kernels.
Depending on the loss function $\ell$, the weights $\alb^t$  can be obtained by using a differentiable kernel method enabling computation of the gradients of the loss w.r.t. the parameters $\thetab$. Such methods include Gaussian Process (GP), Kernel Ridge Regression (KRR), and Logistic Regression (LR).

DKL inherits benefits and drawbacks from deep learning and kernel methods.
It follows that gradient descent algorithms are required to optimize $\thetab$ which can be high dimensional such that seeing a significant amount of training samples is thus essential to avoid overfitting.
However, once the latter condition is met, scalability of the kernel method becomes limiting as the running time of kernel methods scales approximately in $O(m^3)$ for a training set of $m$ samples.
Some approximations of the kernel \citep{williams2001using,wilson2015kernel} are thus needed for the scalability of the DKL method.

\paragraph{Few-Shot DKL:} In the setup of episodic meta learning, also known as few-shot learning, one has access to a meta-training collection $\mathscr{D}_{meta-trn} := \LC (D^{t_j}_{trn}, D^{t_j}_{val}) \RC_{j=1}^{T}$, of $T$ tasks to \textit{learn how to learn} from few datapoints.
Each task $ t_j $ has its own training (or support) set $D^{t_j}_{trn}$ and validation (or query) set $D^{t_j}_{val}$.
A meta-testing collection $\mathscr{D}_{meta-tst}$ is also available to assess the generalization performances of the few-shot algorithm across unseen tasks.
To learn a few-shot DKL method for FSR in such settings, one can share the parameters of $\phib_{\thetab}$ across all tasks, similar to metric learning algorithms.
Hence, for a given task $t_j$, the
inputs are first transformed by the function $\phib_{\thetab}$ and then a kernel method is used to obtain the regressor $h^{t_j}_*$, which will be evaluated on $D^{t_j}_{val}$. \\

\paragraph{KRR:} Using the squared loss and the L2-norm to compute $\norm{h}_{\Rcal}$, KRR gives the optimal regressor for a task $t$ and its validation loss $\Lcal_{\thetab, \rhob, \lambda}^t$ as follows:
\begin{align}\label{eq:krr_optimal}
    h^t_*(\xb) = \alb K_{\xb, trn}, & \quad \textit{ with } \quad \alb = (K_{trn, trn}~+~\lambda I)^{-1}~\yb_{trn}\\
    \Lcal_{\thetab, \rhob, \lambda}^t &=  \esp{\xb,y \sim D^{t}_{val}} (\alb K_{\xb, trn} - y)^2,
\end{align}
where $K_{trn, trn}$ is the matrix of kernel evaluations and each entry $K_{trn, trn:il} = k_{\rhob}(\phib_{\thetab}(\xb_i), \phib_{\thetab}(\xb_l)) $ for $(\xb_i, \cdot), (\xb_l, \cdot) \in D^{t_j}_{trn}$ . An equivalent definition applies to $K_{\xb, trn}$. \\

\paragraph{GP:} If using the negative log likelihood loss function, the GP algorithm gives a probabilistic regressor for the which predictive mean is given by Equation~\ref{eq:krr_optimal} and the loss for a task $t$ is:
\begin{align}
    \Lcal_{\thetab, \rhob, \lambda}^{t} &= - \ln \Ncal ( \yb_{val} ; \mathbb{E}[{h^t_*}], \text{cov}({h^t_*}) ), \\
    \mathbb{E}[{h^t_*}] &= K_{val, trn}(K_{trn, trn} + \lambda I)^{-1}\yb_{trn},  \\
    cov({h^t_*}) &= K_{val, val} - K_{val, trn}(K_{trn, trn} + \lambda I)^{-1} K_{trn, val}
\end{align}

Finally, the parameters $\thetab$ of the neural network, along with $\lambda$ and the hyperparameters $\rhob$ of the kernel, are optimized using the expected loss on all tasks:
\begin{equation} \label{eq:meta_loss_general}
\argmin{\thetab, \rhob, \lambda} \esp{t \sim  \mathscr{D}_{meta-trn}} \Lcal_{\thetab, \rhob, \lambda}^{t}.
\end{equation}
For tractability of this expectation, we use a Monte-Carlo approximation with $b$ tasks. Unless otherwise specified, we use $b=32$ and $m := \vert D^{t_j}_{trn} \vert = 10$, yielding a mini-batch of 320 samples. In our experiments, we have fixed $\lambda$ to $1/|D^{t_j}_{trn}|$.

To summarize, this algorithm finds a representation common to all tasks such that the kernel method (in our case, GP or KRR) will generalize well from a small amount of samples.
Interestingly, this alleviates two of the main limitations of single task DKL. First, the scalability of the kernel method is no longer an issue since we are in the few-shot learning regime\footnote{Even with several hundred samples, the computational cost of embedding each example is usually higher than inverting the Gram matrix.}. Secondly, the parameters $\thetab$ (and $\rhob, \lambda$) are learned across a potentially large amount of tasks and samples, providing the opportunity to learn a complex representation without overfitting.

It is worth mentioning that when using the linear kernel and the KRR algorithm, we recover the few-shot classification algorithm R2-D2 proposed by \citet{bertinetto2018meta}.
Their intent was to show that KRR can be used for fast adaptation at test-time in classification settings as it is differentiable.
In contrast, our intent is to formalize and adapt the DKL framework to FSR and justify how this powerful combination of kernel methods and deep networks can learn covariance functions.

\section{Proposed Method}

\subsection{Adaptive Deep Kernel Method}
\begin{figure}[h]
    \centering
    \includegraphics[width=0.5\textwidth]{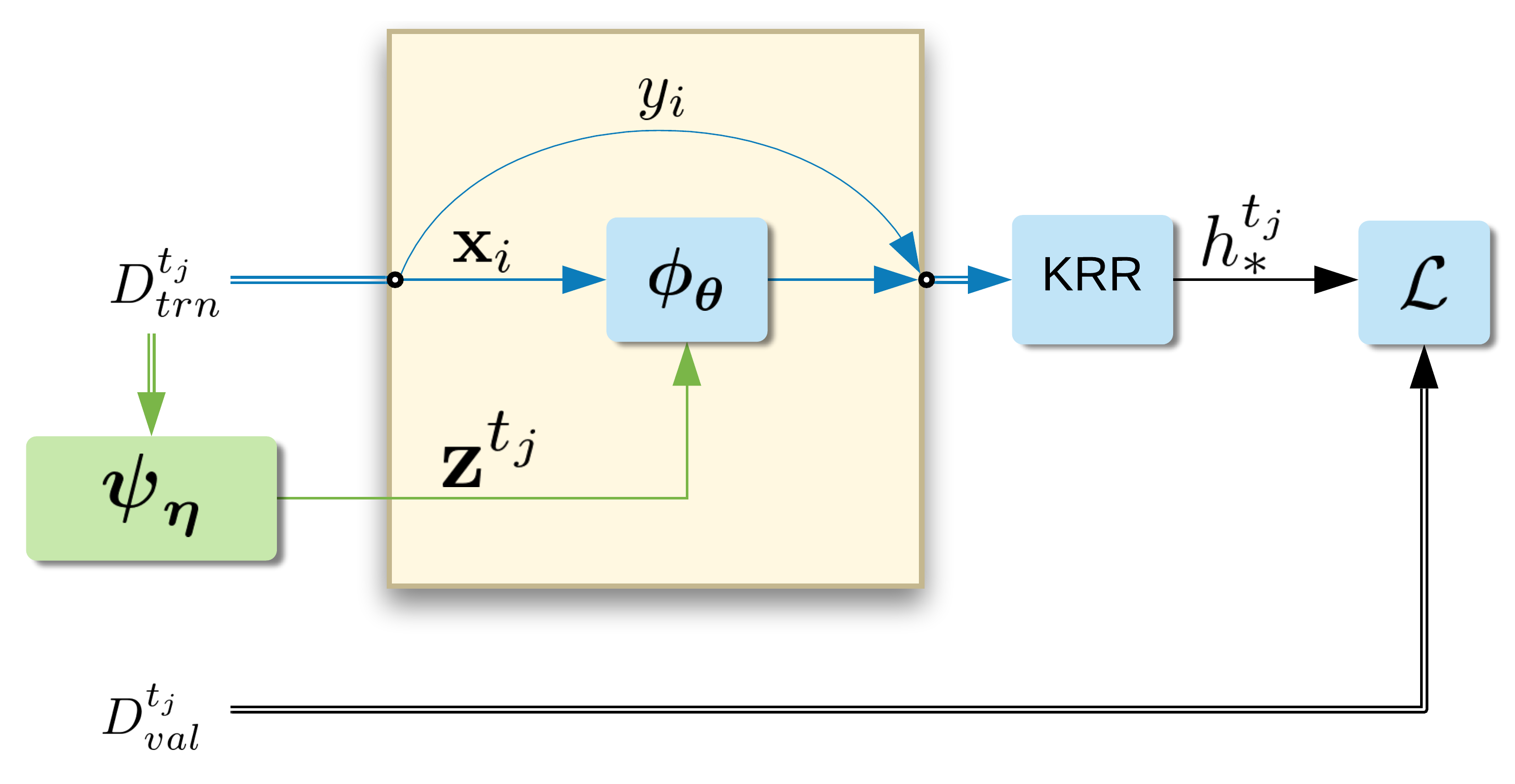}
    \caption{The ADKL algorithm. The green component corresponds to the adaptive part of the algorithm, which contrasts it with standard deep kernel learning approaches.}
    \label{fig:adkl-algorithm}
\end{figure}

As described previously, a DKL algorithm for FSR learns a fixed kernel function $ k_{DKL}(\xb, \xb')~:=~k_{\rhob}(\phib_{\thetab}(\xb), \phib_{\thetab}(\xb')) $ shared across all tasks of interest.
While a regressor on a given task can obtain an arbitrarily small loss as the training set size increases, a fixed kernel might not learn well in the few-shot regime.
To verify this hypothesis, we propose an adaptive deep kernel learning (ADKL) algorithm illustrated in Figure~\ref{fig:adkl-algorithm}.
It learns an adaptive and task-dependent kernel for few-shot learning instead of a single fixed kernel.
We define the adaptive kernel as follows:
\begin{equation}\label{eq:adkl_kernel}
    k_{ADKL}(\xb, \xb' ; \zb_t) := k_{\rhob}(\phib_{\thetab}(\xb ; \zb_t), \phib_{\thetab}(\xb' ; \zb_t))
\end{equation}
where $\zb_t = \psib_{\etab}( D^t_{trn}) $ represents a task embedding obtained by transforming the training set $D^t_{trn}$ with the task encoding network $\psib_{\etab}$.
What follows is a detailed description of the task encoding network $\psib_{\etab}$ and the architecture used to adapt a kernel to a given task $t$.

\paragraph{Task Encoding} One challenge for the network $\psib_{\etab}$  is to capture complex dependencies in the training set $D^t_{trn}$ to provide a useful task encoding $\zb$. Furthermore, the task encoder should be invariant to permutations of the training set and be able to encode a variable amount of samples. After exploring a variety of architectures, we found that more complex ones such as Transformers \cite{vaswani2017attention} tend to underperform.
This is possibly due to overfitting or the challenges associated with training such architectures.

Consequently, we introduce slight modifications to DeepSets, an order invariant network proposed by \citet{zaheer2017deep}.
It begins with the computation of the representation of each input-target pair $\xyb_i =  \rb (\textit{Concat}(\ub(\xb_i), \vb(y_i)) )$ for all  $ (\xb_i, y_i) \in D^t_{trn}$, using neural networks $\ub, \vb, \text{ and } \rb$.
This allows $\xyb_i$  to capture nonlinear interactions between the inputs and the targets if $\rb$ is a nonlinear transformation.
Then, by computing the empirical mean $\mu^t_{\xyb}$ and standard deviation $\sigma^t_{\xyb}$ of the set $\LC \xyb_1, \xyb_2, \ldots, \xyb_m \RC$, we obtain the task representation
\begin{equation}\label{eq:task_repr_deep_set}
\zb^t = \psib_{\etab}( D^t_{trn}) := \wb (\textit{Concat} (\mu^t_{\xyb},  \sigma^t_{\xyb}) )  ,
\end{equation}
where $\wb $ is a also a neural network.
As $\mu^t_{\xyb}$ and $\sigma^t_{\xyb}$ are invariant to permutations in $D^t_{trn}$, it follows that $\psib_{\etab}$ is also permutation invariant.
In summary, $\psib_{\etab}$ is a nonlinear mapping of the first and second moment of the sample representations which were also nonlinear transformations of the original inputs and targets.
The learnable parameters $\etab$ of the task encoder include all the parameters of the networks $\ub, \vb, \rb, \text{ and } \wb$, and are shared across all tasks.

\paragraph{Adapted Kernel Computation} Once the task representation is obtained, we compute the conditional input embedding using the function $\phib_{\thetab}$.
Let $\ub'(\xb)$ be the non-conditional embedding of the input $\xb$ using a neural network $\ub'$, whose parameters are shared with the network $\ub$ within the task encoder.
We simply compute the conditional embedding of inputs as:
\begin{equation}\label{eq:conditonal_embedding}
    \phib_{\thetab}(\xb; \zb_t) = \ob (\textit{Concat} (\ub'(\xb), \zb_t) ) ,
\end{equation}
where $\ob$ is a nonlinear neural network that allows for capturing complex interactions between the task and the input representations. The adapted kernel for a given task is then obtained by combining Equations \ref{eq:adkl_kernel}, \ref{eq:task_repr_deep_set}, \ref{eq:conditonal_embedding}.
The learnable parameters of $\ob$ and  $\ub'$ together constitute $\thetab$ and are shared across all tasks.
Alternatively, different architectures such as Feature-wise Linear Modulation (FiLM) \citep{perez2018film}, Hypernetwork \citep{ha2016hypernetworks}, or Bilinear transformation \citep{tenenbaum2000separating} could be used to compute $\phib_{\thetab}(\xb; \zb)$, though we found that a simple concatenation was sufficient for our applications.

\paragraph{Kernel Methods} To find the regressor that minimizes Equation~\ref{eq:minimal_norm_regression} for each task $t$, we use GP and KRR described in Section \ref{sec:preliminaries}.
In the following, we use ADKL-GP and ADKL-KRR to refer to our algorithm when the task-level regressor is given by GP and KRR, respectively.

\subsection{Meta-Regularization}

\newcommand{\dtrn}{D^{t_j}_{trn}}
\newcommand{\dval}{D^{t_j}_{val}}
\newcommand{\enc}{\psib_{\etab}}

To assist in the training of the ADKL algorithm, we minimize a contrastive loss between $\dtrn$ and $\dval$. Doing so, we  encourage the encoder $\psib_{\etab}$ to embed datasets of the same task closer than those from different tasks.
The non-parametric classification
loss \cite{wu2018unsupervised} and its variants, such as NT-Xent \cite{chen2020simple} and InfoNCE \cite{oord2019representation} are popular choices for the contrastive loss function, but herein we use the InfoNCE.
Using a mini-batch approximation of the expectations, this yields the regularizer $\tilde{I}_{\etab}$ defined as follows:
\begin{align}
    \tilde{I}_{\etab} \eqdef \tfrac{1}{b} \sum_{j=1}^b \enc(\dtrn) \cdot \enc(\dval) - \ln \tfrac{1}{b(b-1)} \sum_{j=1}^b \sum_{i\neq j} e^{\enc(\dtrn) \cdot \enc(D^{t_i}_{val})}.
\end{align}

When adding this to our training objective with $\gamma \geq 0$ as a tradeoff hyperparameter, we have:
\begin{align}
\argmin{\thetab, \etab, \rhob, \lambda} \esp{t_j \sim  B} \Lcal_{\thetab, \etab, \rhob, \lambda}^{t_j}
    - \gamma \tilde{I}_{\etab}.
\end{align}

\section{Related Work}

Our study spans the research areas of deep kernel learning and few-shot learning.
For a comprehensive overview of few-shot learning methods, we refer the reader to \citet{yaqing2019review, chen2019closer}, as we focus on work related to DKL herein.

Across the spectrum of learning approaches, DKL methods lie between neural networks and kernel methods.
While neural networks can learn from a very large amount of data without much prior knowledge, kernel methods learn from fewer data when given an appropriate covariance function that accounts for prior knowledge of the task at hand.
In the first DKL attempt, \citet{wilson2016deep} combined GP with CNN to learn a covariance function adapted to a task from large amounts of data, though the large time and space complexity of kernel methods forced the approximation of the exact kernel using KISS-GP \cite{wilson2015kernel}.
\citet{dasguptafinite} have demonstrated that such approximation is not necessary using finite rank kernels.
Here, we also show that learning from a collection of tasks (FSR mode) does not require any approximation when the covariance function is shared across tasks.
This is an important distinction between our study and other existing studies in DKL, which learn their kernel from single tasks instead of task collections.

On the spectrum between NNs and kernel methods, metric learning also bears mention.
Metric learning algorithms learn an input covariance function shared across tasks but rely only on the expressive power of DNNs.
First, stochastic kernels are built out of shared feature extractors, simple pairwise metrics (e.g. cosine similarity \cite{vinyals2016matching}, Euclidean distance \cite{snell2017prototypical}), or parametric functions (e.g. relation modules \cite{sung2018learning}, graph neural networks \cite{garcia2017few}).
Then, within tasks, the predictions consist of a distance-weighted combination of the training sample labels with the stochastic kernel evaluations---no adaptation is done.
The recently introduced Proto-MAML \cite{triantafillou2019meta} method, which captures the best of Prototypical Networks \cite{snell2017prototypical} and MAML \cite{finn2017model}, allows within-task adaptation using MAML on a network built on top of the kernel function.
Similarly, \citet{kim2018bayesian} have proposed a Bayesian version of MAML where a feature extractor is shared across tasks, while multiple MAML particles are used for the task-level adaptation.
\citet{bertinetto2018meta} have also tackled this lack of adaptation for new tasks by using KRR and Logistic Regression to find the appropriate weighting of the training samples.
This study can be considered the first application of DKL to few-shot learning.
However, its contribution was limited to showing that simple differentiable learning algorithms can increase adaptation in the metric learning framework.
Our work extends beyond by formalizing few-shot learning in the deep kernel learning framework where test-time adaptation is achieved through kernel methods.
We also create another layer of adaptation by allowing task-specific kernels that are created at test-time.

Since ADKL-GP uses GPs, it is related to neural processes \citep{garnelo2018neural}, which proposes a scalable alternative to learning regression functions by performing inference on stochastic processes.
Furthermore, in this family of methods, Conditional Neural Processes (CNP) \citep{garnelo2018conditional} and Attentive Neural Processes (ANP) \citep{kim2019attentive} are even more relevant to our study as both methods learn conditional stochastic processes parameterized by conditions derived from training data points.
While ANP imposes consistency with respect to some prior process, CNP does not and thus does not have the mathematical guarantees associated with stochastic processes.
By comparison, our proposed ADKL-GP algorithm also learns conditional stochastic processes, but within the GP framework, thus benefiting from the associated mathematical guarantees.

\section{Experiments}
\subsection{Datasets}
Previous work on few-shot regression has relied on synthetic datasets to evaluate performance.
We instead introduce two real-world benchmarks drawn from the field of drug discovery.
These benchmarks will allow us to measure the ability of few-shot learning algorithms to adapt in settings where tasks require a considerably different measure of similarity between the inputs.
For instance, when predicting binding affinities between small molecules, the covariance function must learn characteristics of a protein binding site that changes between proteins or tasks.
We describe each dataset used in our experiments below and, unless stated otherwise, the meta-training, meta-validation, and meta-testing contain, respectively, 56.25\%, 18.75\% and 25\% of all of their tasks.
A pre-processed version of these datasets is available with this work \href{https://anonymous.4open.science/r/c1c2daac-6438-41cf-93ef-303322a093cb/}
{here}
\begin{enumerate}
    \item \textbf{Sinusoids}:
     This meta-dataset was recently proposed by \citet{kim2018bayesian} as a challenging few-shot synthetic regression benchmark.
     It consists of 5,000 tasks defined by a sinusoidal functions of the form: $y~=~A~\sin(wx~+~b)~+~\epsilon$.
     The parameters $A, w, b$ characterize each task and are drawn from the following intervals: $A \in \LB0.1, 5.0 \RB$, $b \in \LB 0.0, 2\pi \RB$, $w \in \LB0.5, 2.0\RB $.
     Samples for each task are generated by sampling inputs $ x \in [-5.0, 5.0] $ and  observational noise from $ \epsilon \sim N (0,(0.01A)^2) $.
     Every model on this task collection uses a fully connected network of 2 layers of 120 hidden units as its input feature extractor.
     Visuals of ground truths and predictions for some functions are shown in \ref{app-pred-curves}.

    \item \textbf{Binding}:
    The goal here is to predict the binding affinity of small, drug-like molecules to proteins.
    This task collection was extracted from the public database BindingDB\footnote{Original data available at \url{www.bindingdb.org}} and encompasses 7,620 tasks, each containing between 7 and 9,000 samples.
    Each task corresponds to a protein sequence, which thus defines a separate distribution on the input space of molecules and the output space of (real-valued) binding readouts.

    \item \textbf{Antibacterial}:
    The goal here is to predict the antimicrobial activity of small molecules for various bacteria.
    The task collection was extracted from the public database PubChem\footnote{Available at \url{https://pubchem.ncbi.nlm.nih.gov/}} and contains 3,842 tasks, each consisting of 5 to 225 samples.
    A task corresponds to a combination of a bacterial strain and an experimental setting, which define different data distributions.
\end{enumerate}

For both real-world datasets, the molecules are represented by their SMILES\footnote{See \url{https://en.wikipedia.org/wiki/Simplified_molecular-input_line-entry_system}} encoding, which are descriptions of the molecular structure using short ASCII strings. All models evaluated on these collections share the same input feature extractor configuration: a 1-D CNN of 2 layers of 128 hidden units each and a kernel size of ~5.
We choose CNN instead of LSTM or advanced graph convolutions methods for scalability reasons. Finally, all targets were scaled linearly between 0 and 1.

\subsection{Benchmarking analysis}
We evaluate model performance against R2-D2 \cite{bertinetto2018meta}, CNP\cite{garnelo2018conditional}, and MAML\cite{finn2017model}.
Comparing R2-D2 to ADKL-KRR when using the linear kernel allows us to determine whether the adapted deep kernel provides improved test-time adaptation.
CNP is also a relevant comparison to ADKL-GP and allows us to measure performance differences between the task-level Bayesian models generated within the GP and CNP frameworks.
MAML is considered herein for its fast-adaptation at test-time and as the representative of initialization and optimization based methods.
In the following experiments, all DKL methods use the linear kernel.

\begin{figure}[ht]
  \centering
  \begin{subfigure}{.32\textwidth}
    \includegraphics[width=\textwidth]{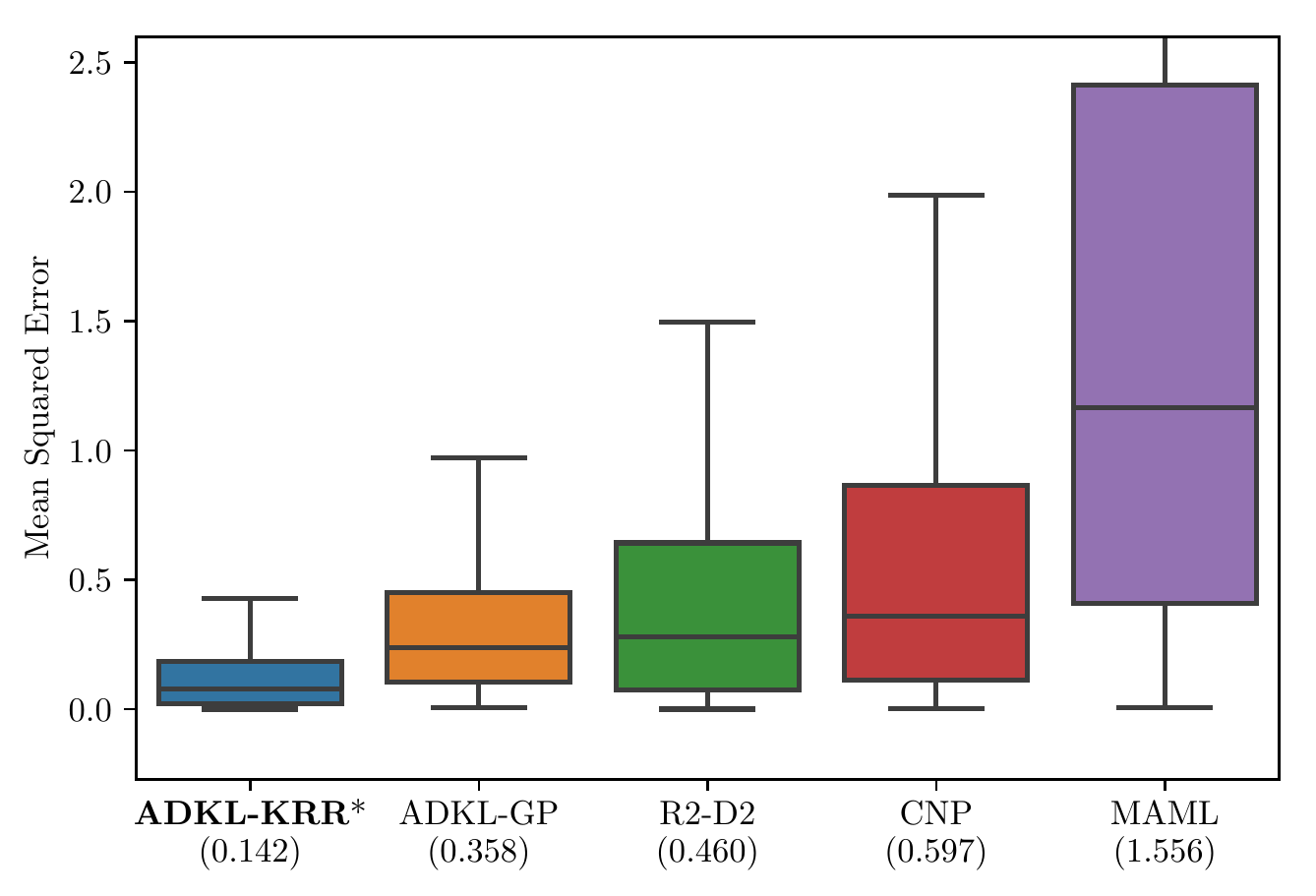}
    \caption{\texttt{Sinusoids} collection}
  \end{subfigure}
  \hfill
  \begin{subfigure}{.32\textwidth}
    \includegraphics[width=\textwidth]{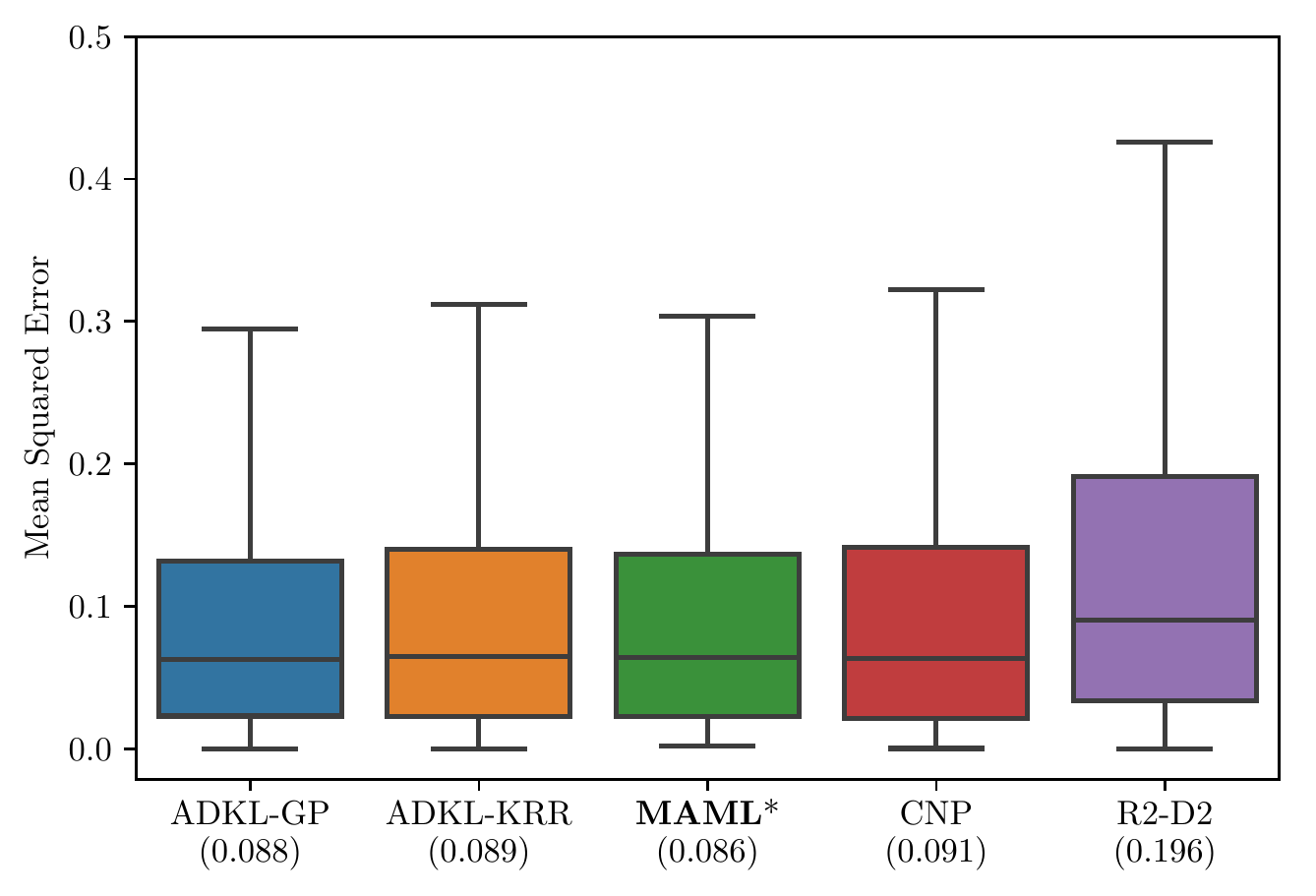}
    \caption{\texttt{Binding} collection}
  \end{subfigure}
  \hfill
  \begin{subfigure}{.32\textwidth}
    \includegraphics[width=\textwidth]{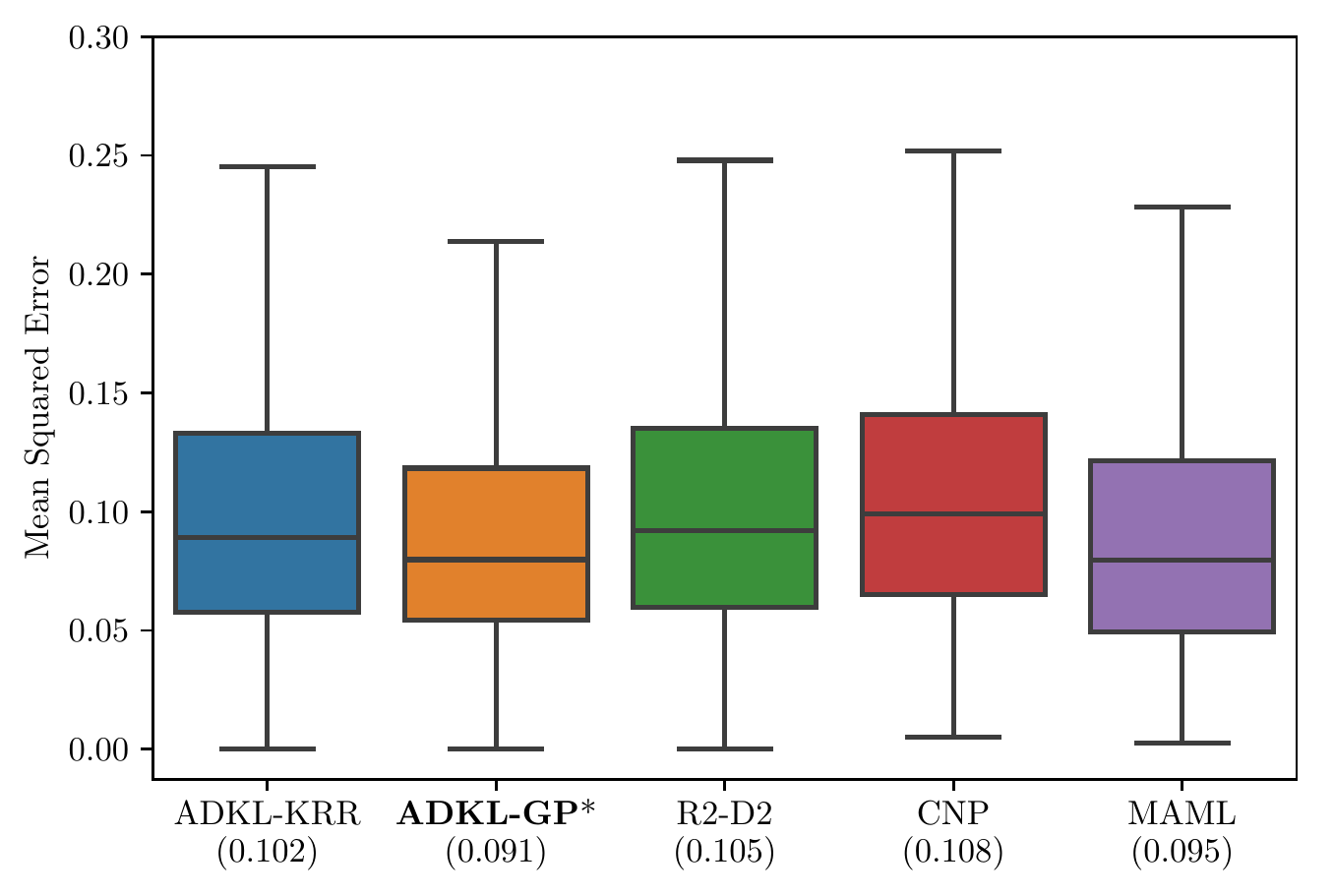}
    \caption{\texttt{Antibacterial} collection}
  \end{subfigure}
  \caption{MSE distributions for tasks in the meta-testing sets. Only the best model for each dataset is marked with an asterisk and the mean MSE for the whole meta-test is shown below the algorithm name.}
  \label{fig:bcomp}
\end{figure}

Our first set of experiments evaluates performance on both the real-world and toy tasks.
We train each method using support and query sets of size $m~=~10$.
During meta-testing, the support set size is also $m~=~10$, but the query set consists of the remaining samples of each task.
For the datasets lacking sufficient samples such as in the Binding and Antibacterial collections, we use half of the samples in the support set and the remaining in the query set.
During meta-testing of each task, we average the Mean Squared Error (MSE) over 20 random partitions of the query and support sets.
We refer to this value as the task MSE.
Figure~\ref{fig:bcomp} illustrates the task MSE distributions over tasks for each collection and algorithm (best hyperparameters were chosen on the meta-validation set).
In general, we observe that the real-world datasets are challenging for all methods but ADKL methods consistently outperform R2-D2 and CNP.
The MSE difference between ADKL-KRR and R2-D2 support the importance of adapting the kernel to each task rather than sharing a single kernel.
Furthermore, we observe that ADKL-GP outperforms CNP, showing the effectiveness of the ADKL approach in comparison with conditional neural processes.
Finally, both adaptive deep kernel methods (ADKL-KRR and ADKL-GP) seem to perform comparably, despite different objective functions.

\begin{figure}[ht]
  \centering
  \begin{subfigure}{.45\textwidth}
    \includegraphics[width=0.8\textwidth, height=3cm]{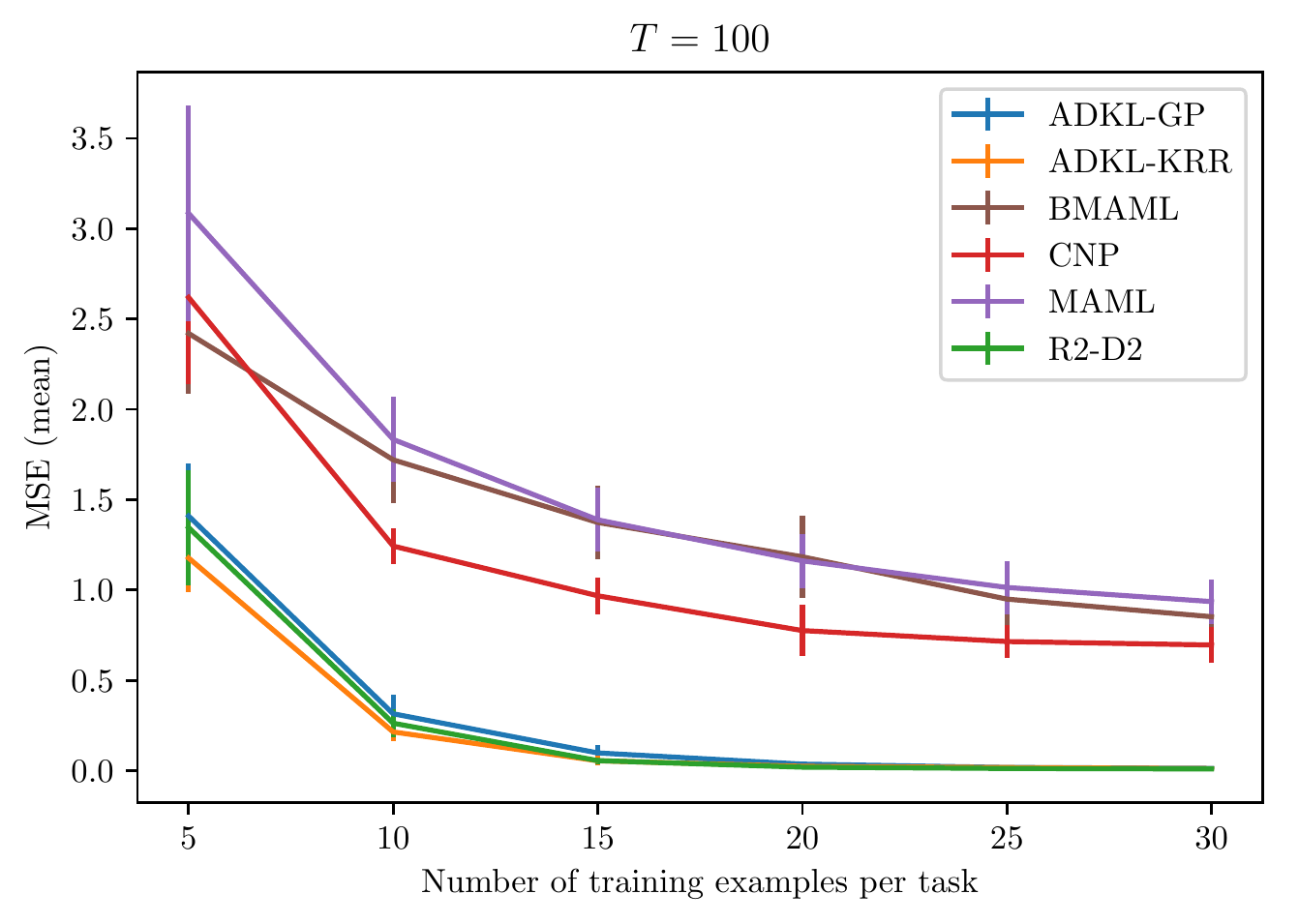}
  \end{subfigure}
  \begin{subfigure}{.45\textwidth}
    \includegraphics[width=0.8\textwidth, height=3cm]{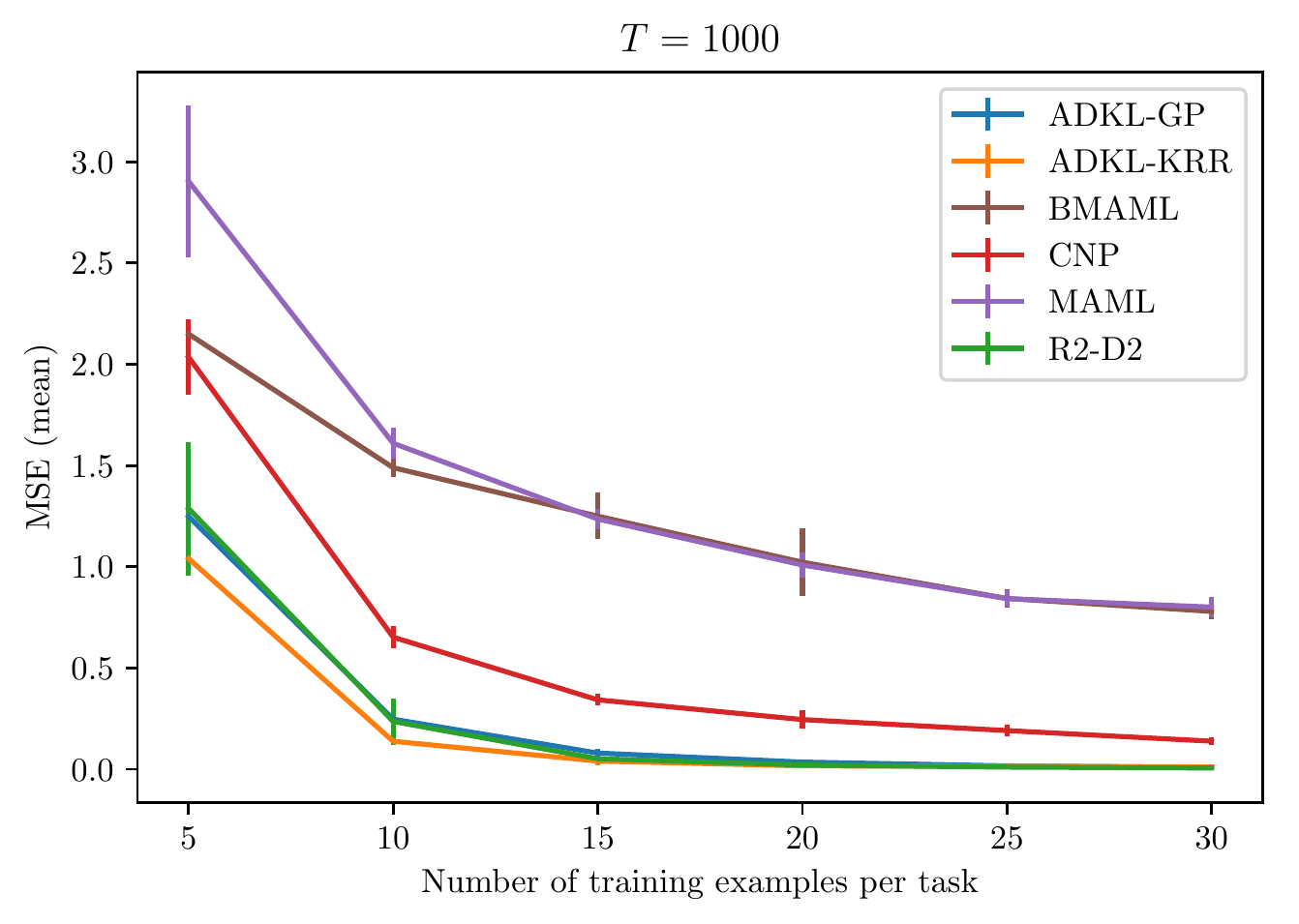}
  \end{subfigure}
  \caption{Average MSE over the testing tasks of the Sinusoids collection. The error bars represent the uncertainty on the average over 5 runs with different seeds.}
  \label{fig:bcomp_by_k}
\end{figure}

A second experimental set is used to measure the across-task generalization and within-task adaptation capabilities of the different methods.
We do so by controlling the number of training tasks ($T \in \LC 100, 1000 \RC $) and the size of the support sets ($m \in \LC 5, 10, 15, 20, 25, 30 \RC $) during meta-training and meta-testing.
Only the Sinusoids collection was used, as experiments with the real-world collections were deemed too time-consuming for the scope of this study.
One would expect the algorithms to generalize poorly to new tasks for lower values of $T$, and their task-level models to adapt poorly to new samples for small values of~$m$.
However, as illustrated in Figure~\ref{fig:bcomp_by_k}, all DKL methods generalize better across tasks than other algorithms.
DKL methods also demonstrate improved within-task generalization using as few as 15 samples, while other methods require more samples to achieve similar MSE.
Moreover, for small support sets, ADKL-KRR shows better within-task generalization than ADKL-GP and R2-D2.
The difference in performance between ADKL-KRR and R2-D2 can be attributed to the kernel adaptation at test-time as this is the only difference between both methods.
When $m$ is small, the difference between ADKL-GP and ADKL-KRR can be attributed to larger predictive uncertainty in GP as the number of samples gets smaller.

\subsection{Active Learning}
What follows aims to measure the effectiveness of the uncertainty captured by the predictive distribution of ADKL-GP for active learning.
By comparing to CNP, we'll quantify which of CNP or GP better captures the data uncertainty for improving FSR under active sample selection.
For this purpose, we meta-train both algorithms using support and query sets of size $m~ =~5$  and use $T \in \LC 100, 1000\RC$ for the Sinusoids collection.
During meta-test time, five samples are randomly selected to constitute the support set $D_{trn}$ and build the initial hypothesis for each task.
Then, from a pool $U$ of unlabeled data, we choose the input $\xb^*$ of maximum predictive entropy, i.e., $\xb^* = \mathrm{argmax}_{\xb \in U}{\mathbb{E} \LB \log p(y \lvert \xb, D_{trn}) \RB}$.
The latter example is removed from $U$ and added to $D_{trn}$ with its predicted label.
The within-task adaptation is performed on the new support set to obtain a new hypothesis which is evaluated on the query set $D_{val}$ of the task.
This process is repeated until we reach the allowed budget of $20$ queries.

\begin{figure}[ht]
  \centering
  \begin{subfigure}{.49\textwidth}
    \includegraphics[width=\textwidth]{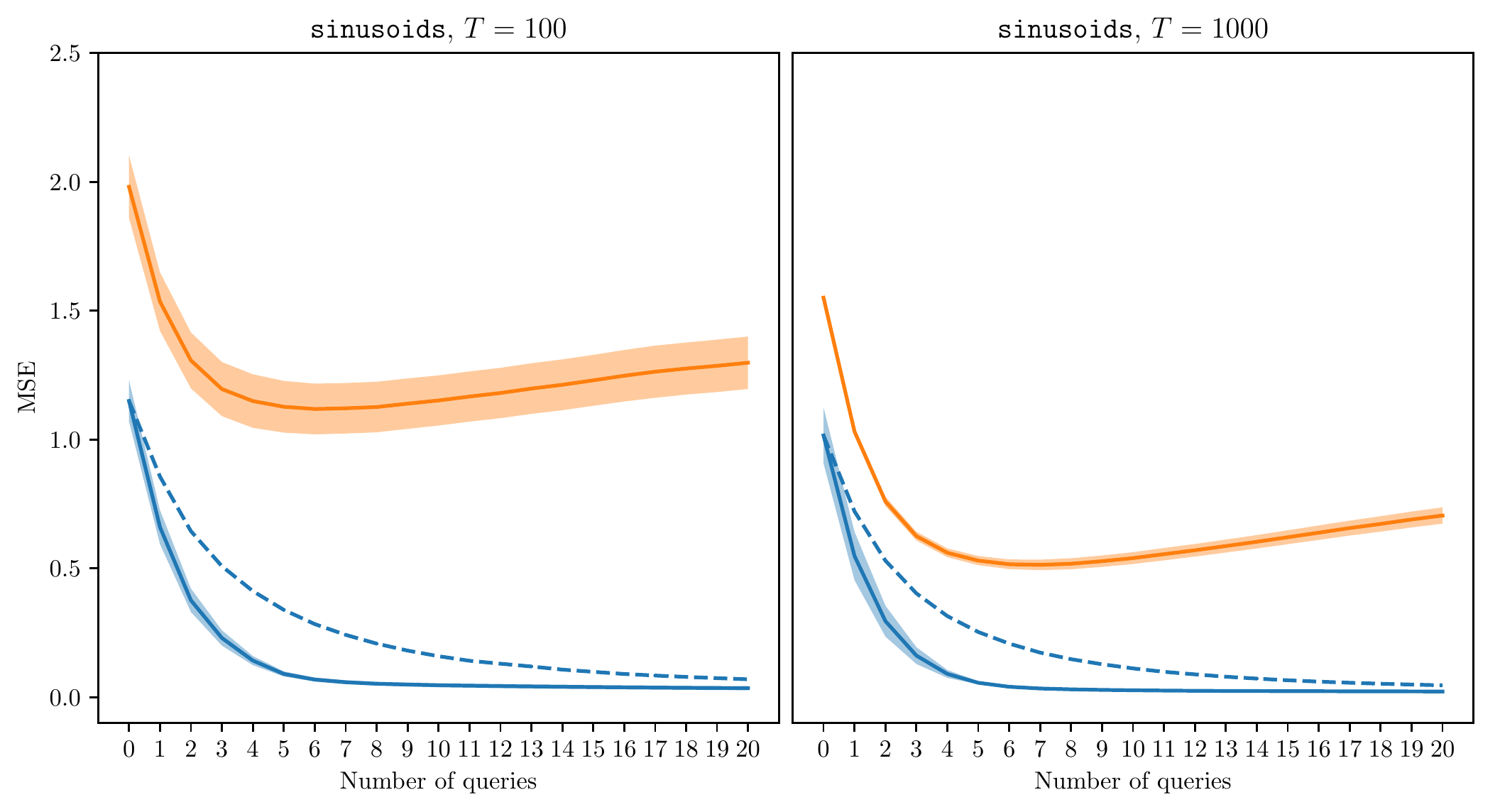}
  \end{subfigure}
    \hfill
  \begin{subfigure}{.49\textwidth}
    \includegraphics[width=\textwidth]{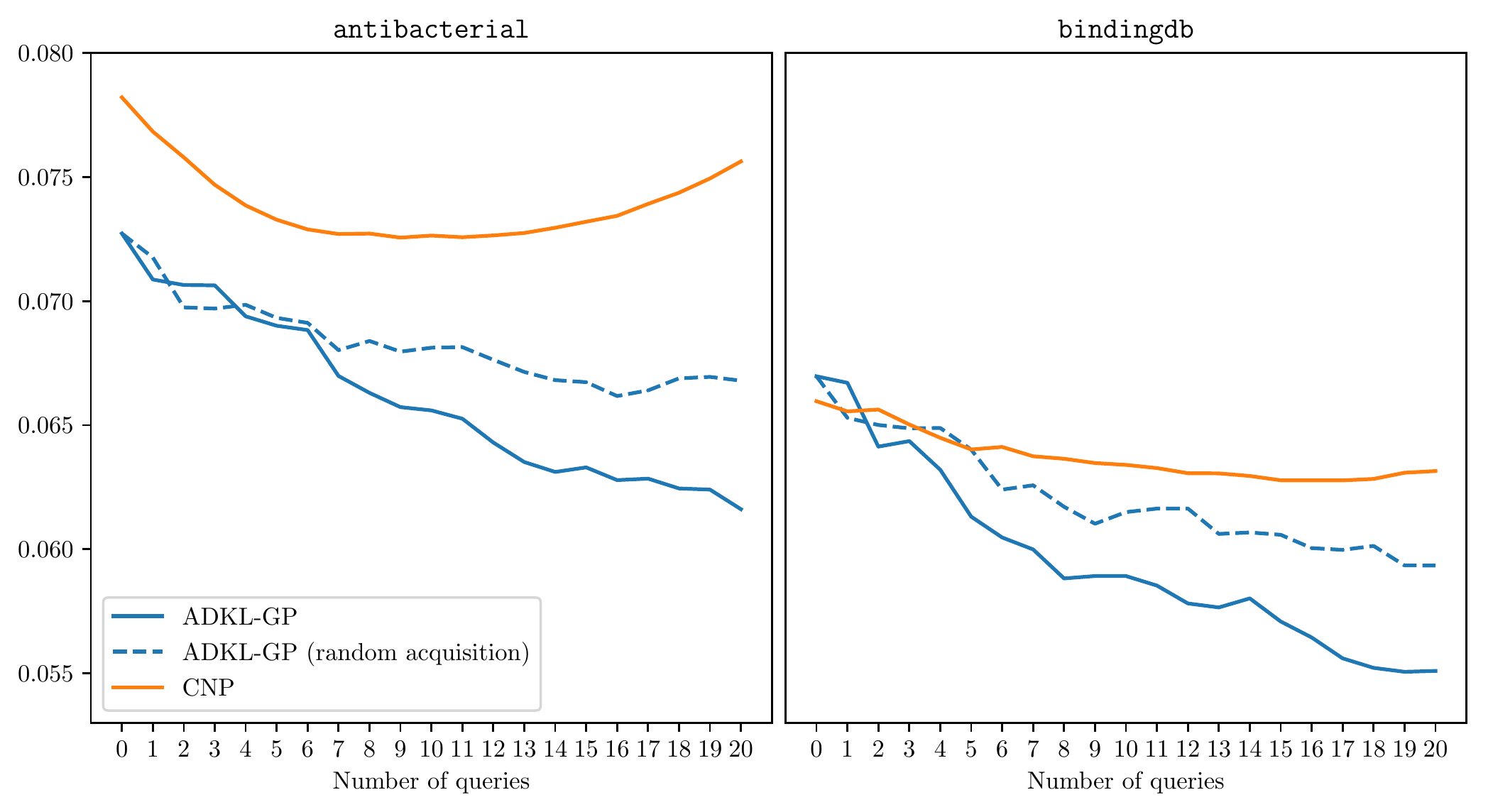}
  \end{subfigure}
  \caption{Average MSE performance on the meta-test during active learning. The width of the shaded regions denotes the uncertainty over five runs for the \texttt{sinusoidal} collection. No uncertainty is shown for the real-world tasks as they were too time consuming.}
  \label{fig:active_ml}
\end{figure}

Figure~\ref{fig:active_ml} highlights that, in the active learning setting, ADKL-GP consistently outperforms CNP.
Very few samples are queried by ADKL-GP to appropriately capture the data distribution.
In contrast, CNP performance is far from optimal, even when allowed the maximum number of queries.
Also, since using the maximum predictive entropy strategy is better than querying samples at random for ADKL-GP (solid vs. dashed line), these results suggest that the predictive uncertainty obtained with GP is more informative and accurate than that of CNP.
Moreover, when the number of queries is greater than $10$, we observe a performance degradation for CNP, while ADKL-GP remains consistent.
This observation highlights the generalization capacity of DKL methods, even outside the few-shot regime where they have been trained.
We attribute this property of DKL methods to their use of kernel methods.

\subsection{Impact of Regularization and Kernel}
\begin{figure}[ht]
  \centering
  \begin{subfigure}{.45\textwidth}
    \includegraphics[width=\textwidth]{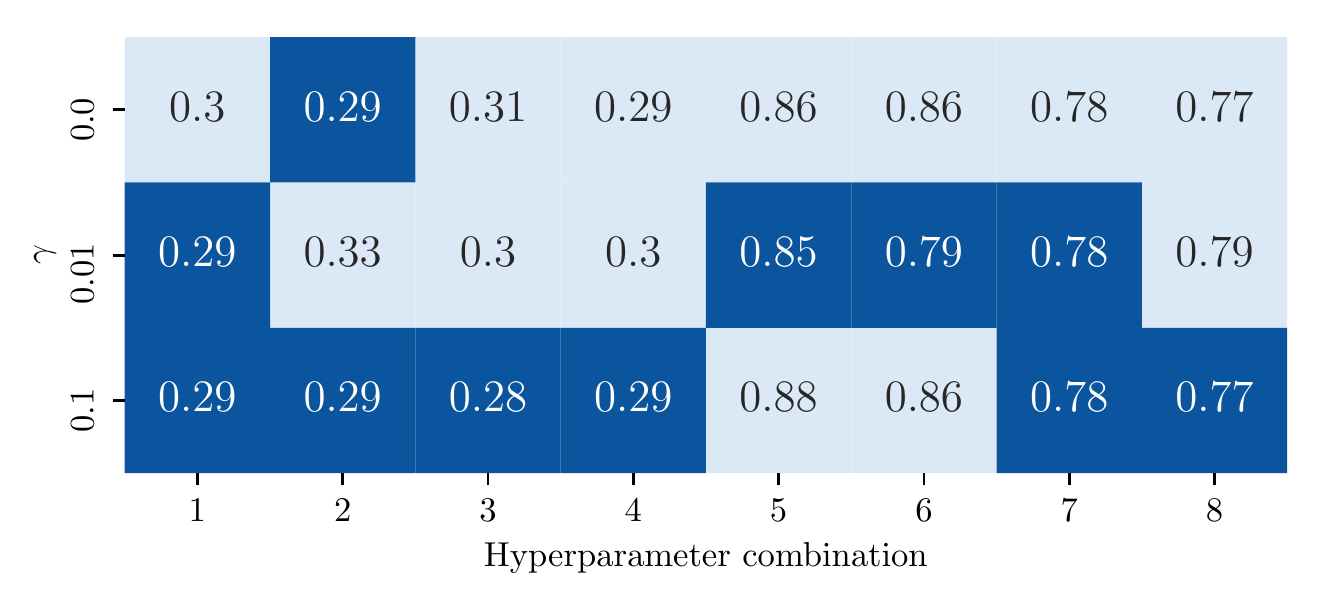}
  \end{subfigure}
  \hfill
  \begin{subfigure}{.54\textwidth}
    \includegraphics[width=\textwidth]{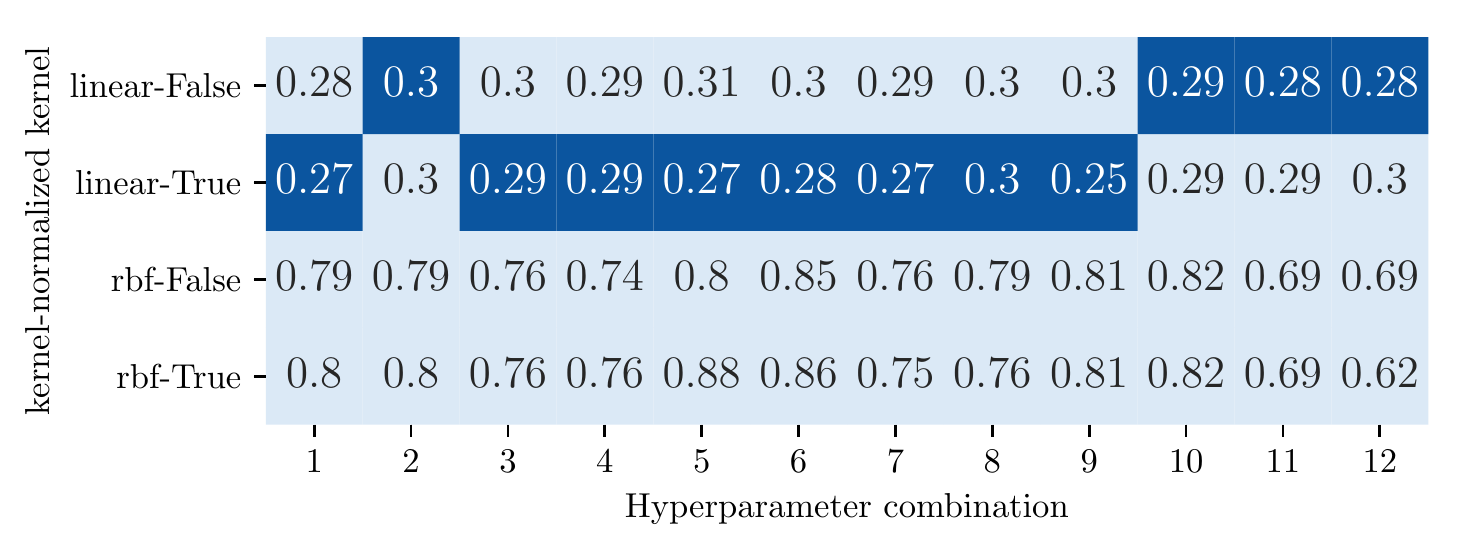}
  \end{subfigure}
  \caption{Impact of $\gamma$ and the kernel on the average meta-test MSE for different hyperparameter combinations (dark blue is best). The combinations are detailed in the \ref{app-kernel-impact} and \ref{app-reg-impact}}
  \label{fig:ablation}
\end{figure}

In our final set of experiments, we take a closer look at the impact of the base kernel and the meta-regularization factor on the generalization during meta-testing.
We do so by evaluating ADKL-KRR on the Sinusoids collection with different hyperparameter combinations.
Figure~\ref{fig:ablation}~(left) shows the performances obtained by varying the meta-regularization parameter $\gamma$ over different hyperparameter configurations (listed in \ref{app-reg-impact}).
When looking at values of $ \gamma \in \LC 0, 0.1, 0.01 \RC$, we observe that non-zero values help in most cases, meaning that the added regularizer improves slightly the task encoder learning.
As choosing the appropriate kernel function is crucial for kernel methods, we also measure how different base kernels impact the learning process.
To do so, we test the linear and RBF kernels, and their normalized versions as given by $ k'(\xb, \xb') := {k(\xb, \xb')} / {\sqrt{k(\xb, \xb) k(\xb', \xb')}} $.
Over different hyperparameter combinations, we observe that the linear kernel yields better generalization performances than the RBF kernel (see Figure~\ref{fig:ablation}~right).
Such a result is within expectation as the scaling parameter of the RBF kernel is shared across tasks, making it more difficult to adapt the deep kernel.
In future work, it would be interesting to explore learning the base kernel hyperparameters using a network similar to $\psib $.

\section{Conclusion}
In this paper, we investigated the benefits of deep kernel learning (DKL) methods for few-shot regression (FSR).
By comparing methods on real-world and synthetic task collections, we successfully demonstrated the effectiveness of the DKL framework in FSR.
Both ADKL-GP and ADKL-KRR outperform the single kernel DKL method, providing evidence of their improved adaptation capacity at test-time through adaptation of the kernel.
Given the Bayesian nature of ADKL-GP, this variation of the main algorithm allows for improvement of the learned models at test-time, providing unique opportunities in settings such as drug discovery where active learning can be deployed in design-make-test cycles.
Finally, by making our drug discovery task collections publicly available, we hope to encourage the machine learning community to work toward new FSR algorithms that can be better adapted to real-world drug-discovery settings.





\newpage

\appendix
\gdef\thesection{Appendix \Alph{section}:}

\section{Regularization impact}\label{app-reg-impact}

Table \ref{tab:ablation-joint-training} presents the hyperparameter combinations used in the experiments to assess the impact of the joint training parameter $\gamma$.

\begin{table}[ht]
    \caption{Effect of using joint training (parameter $\gamma$) on the MSE performances}
    \centering
    \begin{tabular}{llllrrr}
        \toprule
          &     &    & $\gamma$ &    0.0 &   0.01 &    0.1 \\
        Run & kernel & hidden size (target FE) & hidden size (input FE) &        &        &        \\
        \midrule
        1 & linear & [32, 32] & [128, 128, 128] &  0.297 &  \textbf{0.290} &  \textbf{0.290} \\
        2 & linear & [32, 32] & [128, 128] &  \textbf{0.288} &  0.328 &  \textbf{0.288} \\
        3 & linear & [] & [128, 128, 128] &  0.306 &  0.300 &  \textbf{0.285} \\
        4 & linear & [] & [128, 128] &  0.290 &  0.297 &  \textbf{0.287} \\
        5 & rbf & [32, 32] & [128, 128, 128] &  0.864 &  \textbf{0.851} &  0.876 \\
        6 & rbf & [32, 32] & [128, 128] &  0.859 &  \textbf{0.787} &  0.864 \\
        7 & rbf & [] & [128, 128, 128] &  0.780 &  \textbf{0.778} &  \textbf{0.778} \\
        8 & rbf & [] & [128, 128] &  0.769 &  0.793 &  \textbf{0.768} \\
        \bottomrule
    \end{tabular}
    \label{tab:ablation-joint-training}
\end{table}

Note that the performance is significantly worse when using RBF kernel.

\section{Kernel impact}\label{app-kernel-impact}

Table \ref{tab:ablation-kernel} shows the hyperparameter combinations we used to assess the effect of using different kernels, as well as the impact of normalizing them.

\begin{table}[ht]
    \centering
    \caption{MSE performance depending on the kernel choice and normalization}
    \begin{tabular}{llllrrrr}
        \toprule
           &     &    & kernel & \multicolumn{2}{l}{Linear} & \multicolumn{2}{l}{RBF} \\
           &     &    & normalized &  False &   True &  False &   True \\
        Run & encoder arch & target FE & input FE &        &        &        &        \\
        \midrule
        1  & CNP & [32, 32] & [128, 128, 128] &  0.279 &  \textbf{0.274} &  0.786 &  0.795 \\
        2  & CNP & [32, 32] & [128, 128] &  \textbf{0.299} &  0.300 &  0.789 &  0.799 \\
        3  & CNP & [] & [128, 128, 128] &  0.304 &  \textbf{0.289} &  0.761 &  0.755 \\
        4  & CNP & [] & [128, 128] &  0.295 &  \textbf{0.293} &  0.742 &  0.757 \\
        5  & DeepSet & [32, 32] & [128, 128, 128] &  0.313 &  \textbf{0.269} &  0.804 &  0.877 \\
        6  & DeepSet & [32, 32] & [128, 128] &  0.301 &  \textbf{0.277} &  0.849 &  0.856 \\
        7  & DeepSet & [] & [128, 128, 128] &  0.292 &  \textbf{0.273} &  0.764 &  0.754 \\
        8  & DeepSet & [] & [128, 128] &  0.303 &  \textbf{0.298} &  0.788 &  0.763 \\
        9  & KRR & [32, 32] & [128, 128, 128] &  0.296 &  \textbf{0.246} &  0.815 &  0.815 \\
        10 & KRR & [32, 32] & [128, 128] &  \textbf{0.289} &  0.290 &  0.824 &  0.824 \\
        11 & KRR & [] & [128, 128, 128] &  \textbf{0.279} &  0.291 &  0.690 &  0.694 \\
        12 & KRR & [] & [128, 128] &  \textbf{0.275} &  0.299 &  0.685 &  0.622 \\
        \bottomrule
    \end{tabular}
    \label{tab:ablation-kernel}
\end{table}

\section{Prediction curves on the Sinusoids collection}\label{app-pred-curves}

\begin{figure}[h]
    \centering
    \begin{subfigure}{.32\textwidth}
        \centering
        \includegraphics[width=\textwidth]{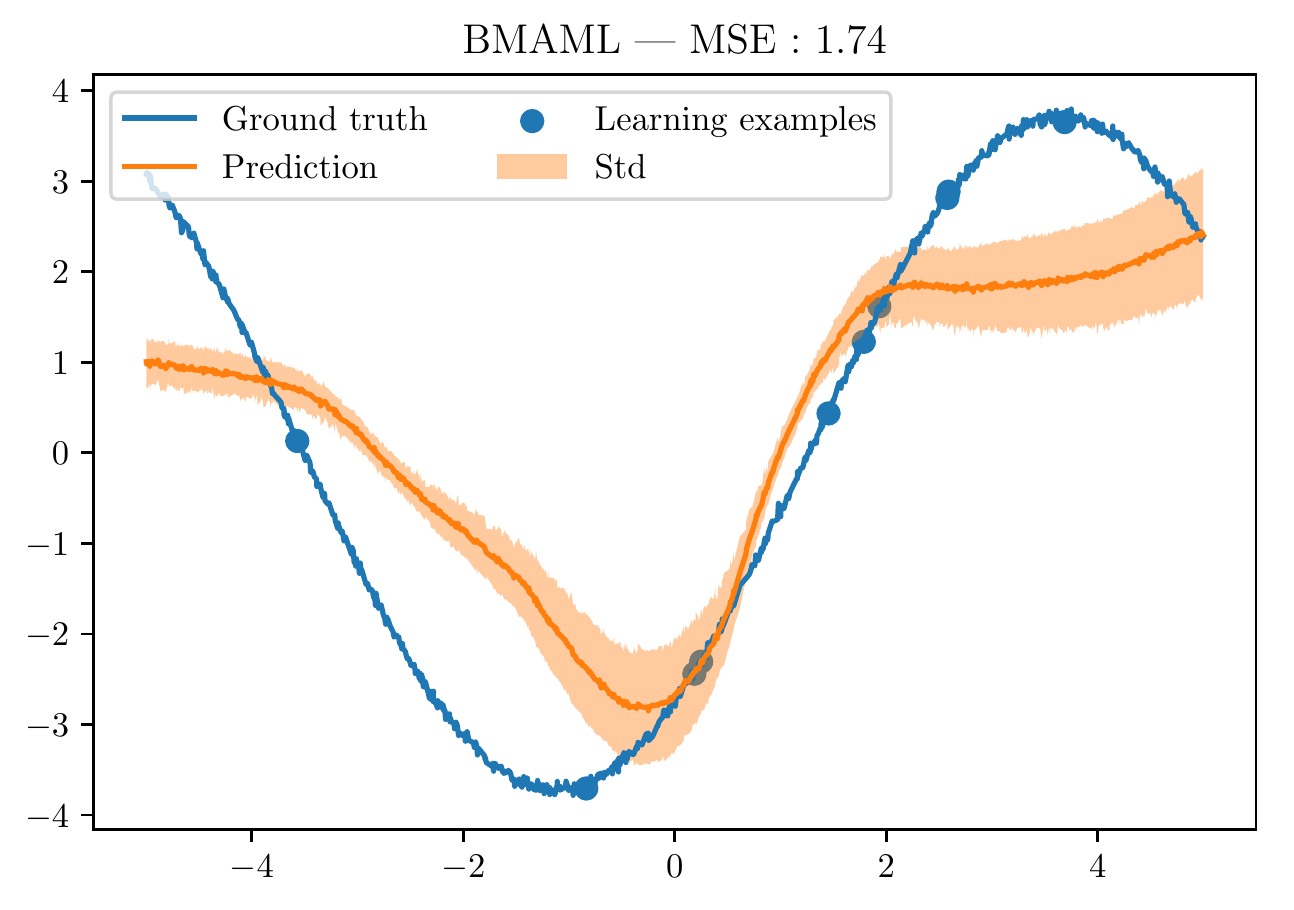}
    \end{subfigure}
    \hfill
    \begin{subfigure}{.32\textwidth}
        \centering
        \includegraphics[width=\textwidth]{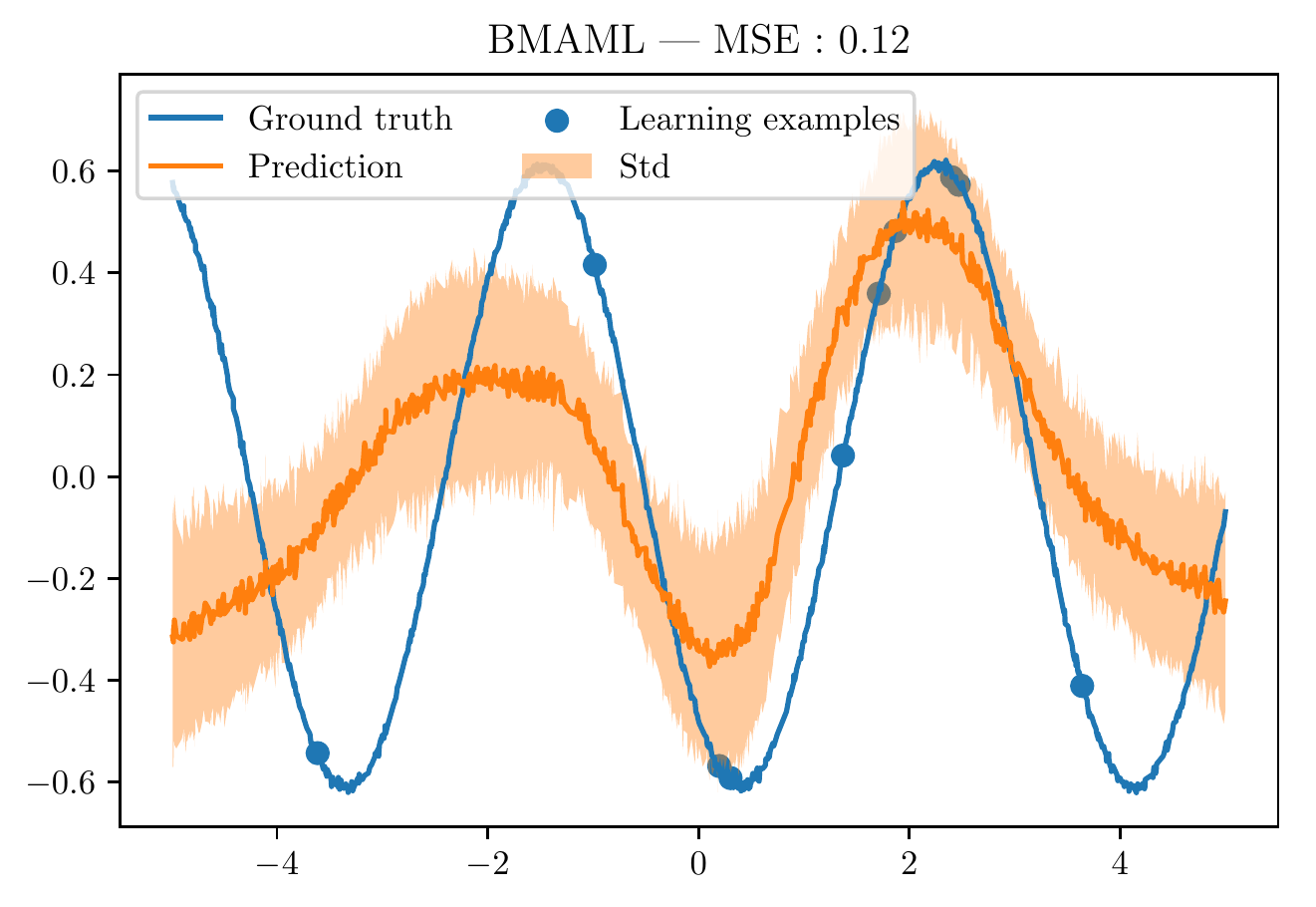}
    \end{subfigure}
    \hfill
    \begin{subfigure}{.32\textwidth}
        \centering
        \includegraphics[width=\textwidth]{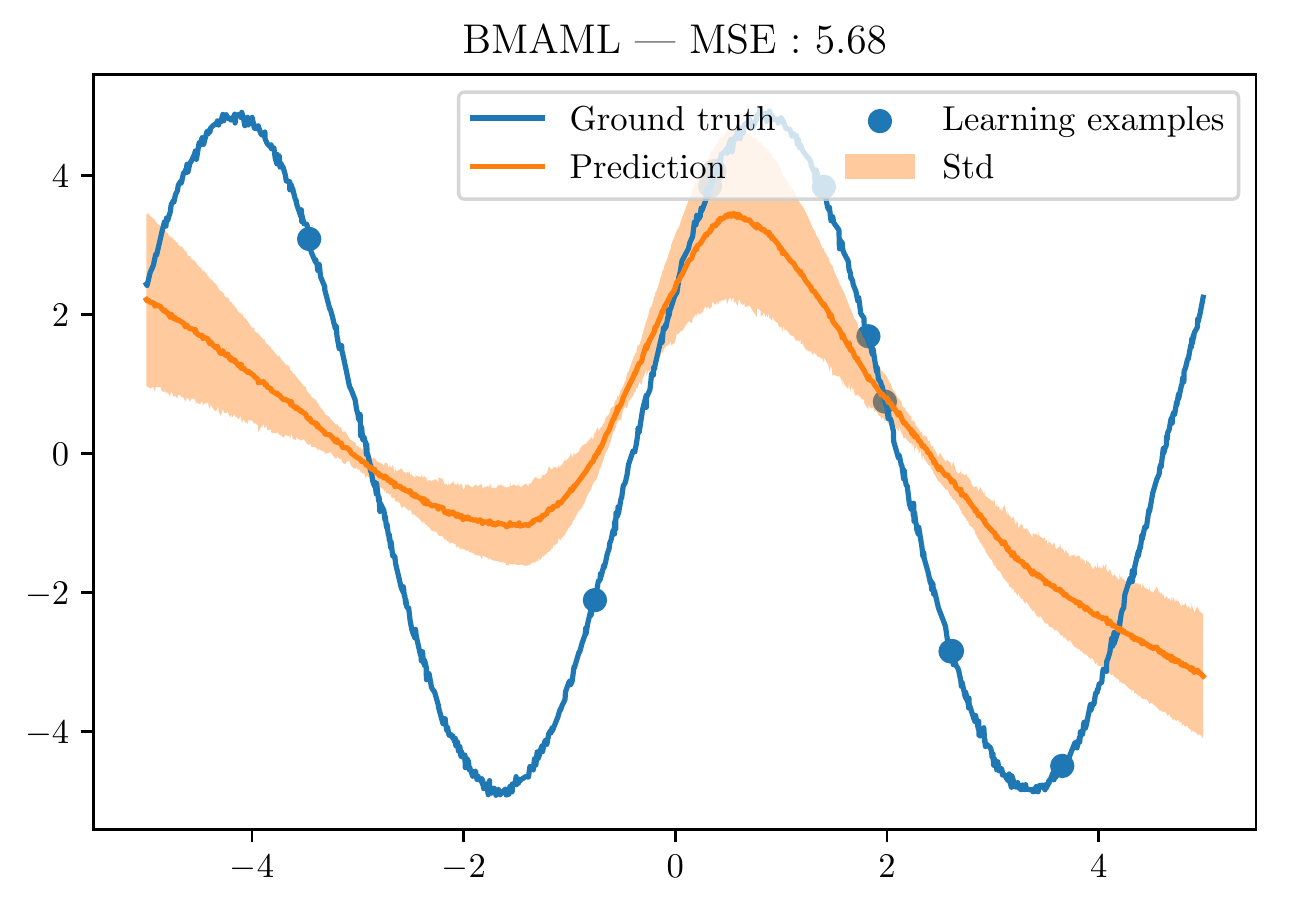}
    \end{subfigure}
    \\
    \begin{subfigure}{.32\textwidth}
        \centering
        \includegraphics[width=\textwidth]{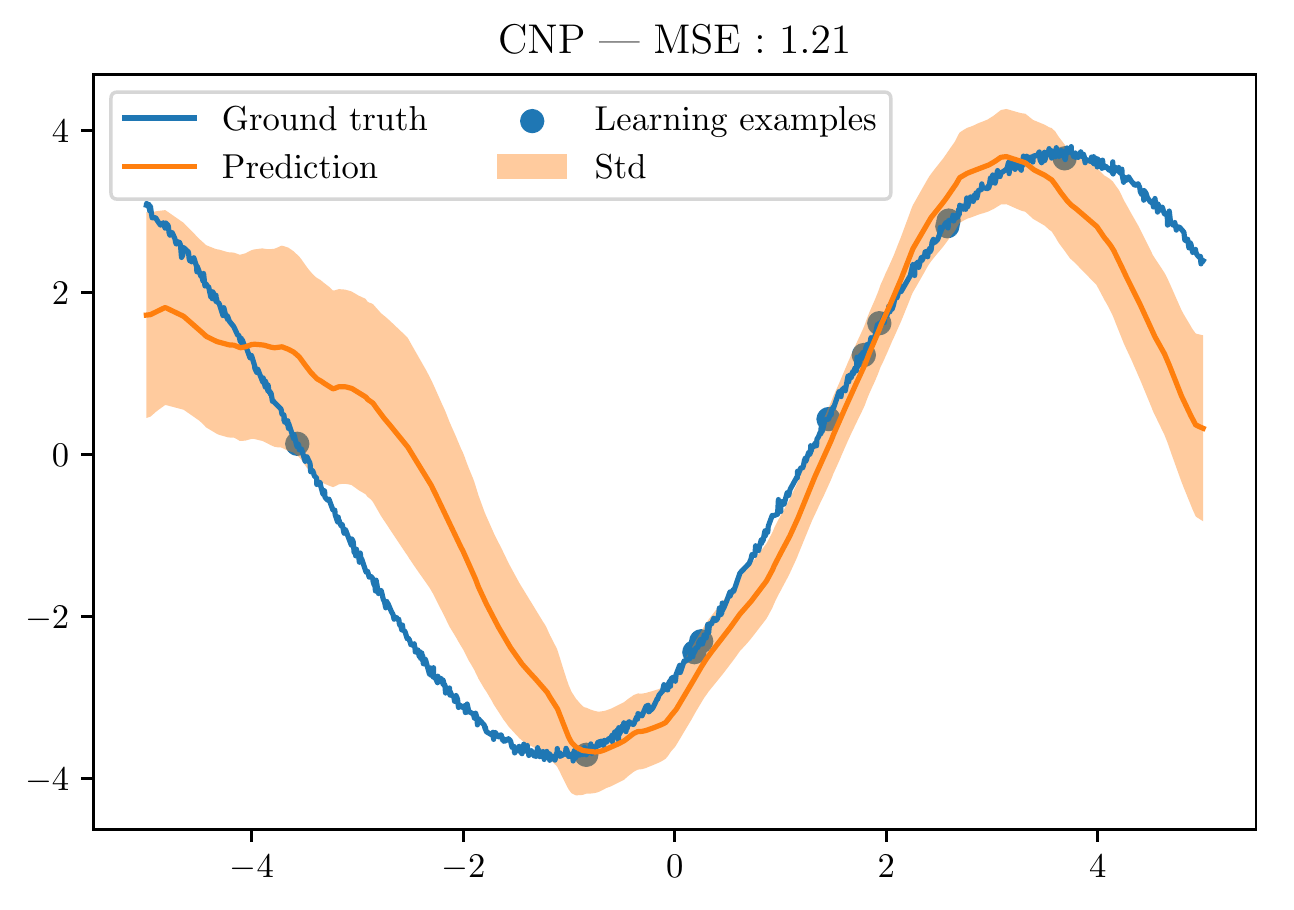}
    \end{subfigure}
    \hfill
    \begin{subfigure}{.32\textwidth}
        \centering
        \includegraphics[width=\textwidth]{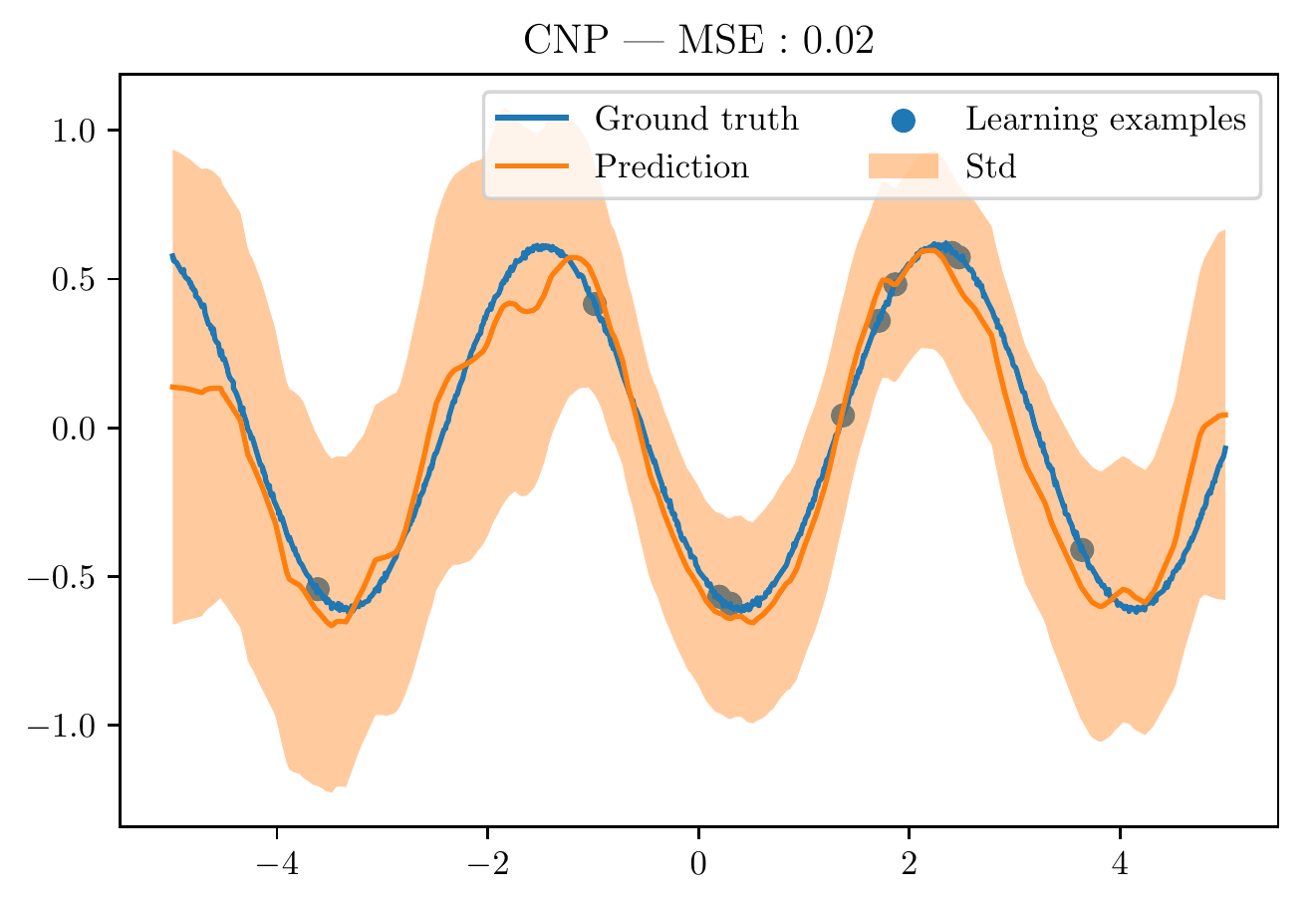}
    \end{subfigure}
    \hfill
    \begin{subfigure}{.32\textwidth}
        \centering
        \includegraphics[width=\textwidth]{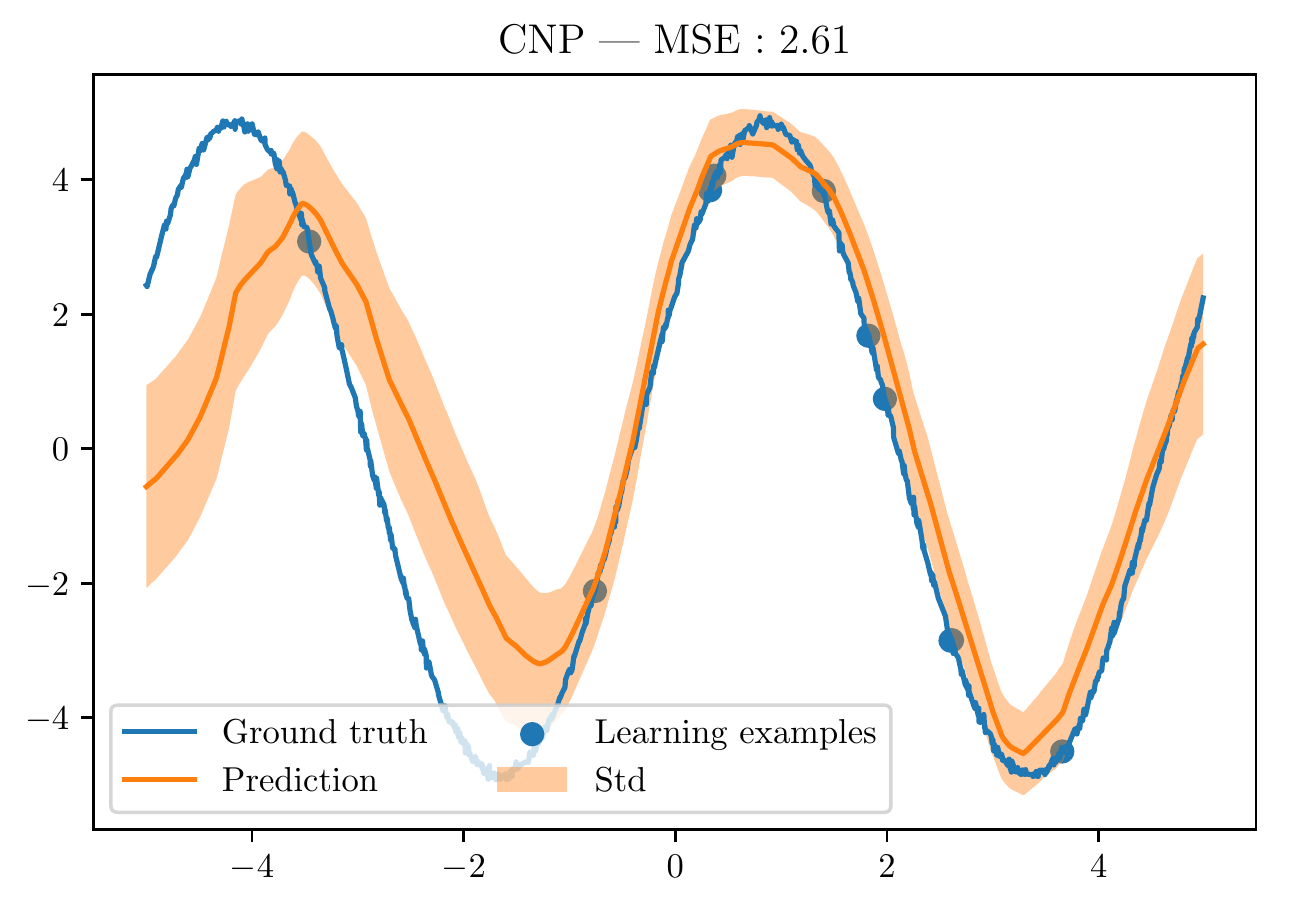}
    \end{subfigure}
    \\
    \begin{subfigure}{.32\textwidth}
        \centering
        \includegraphics[width=\textwidth]{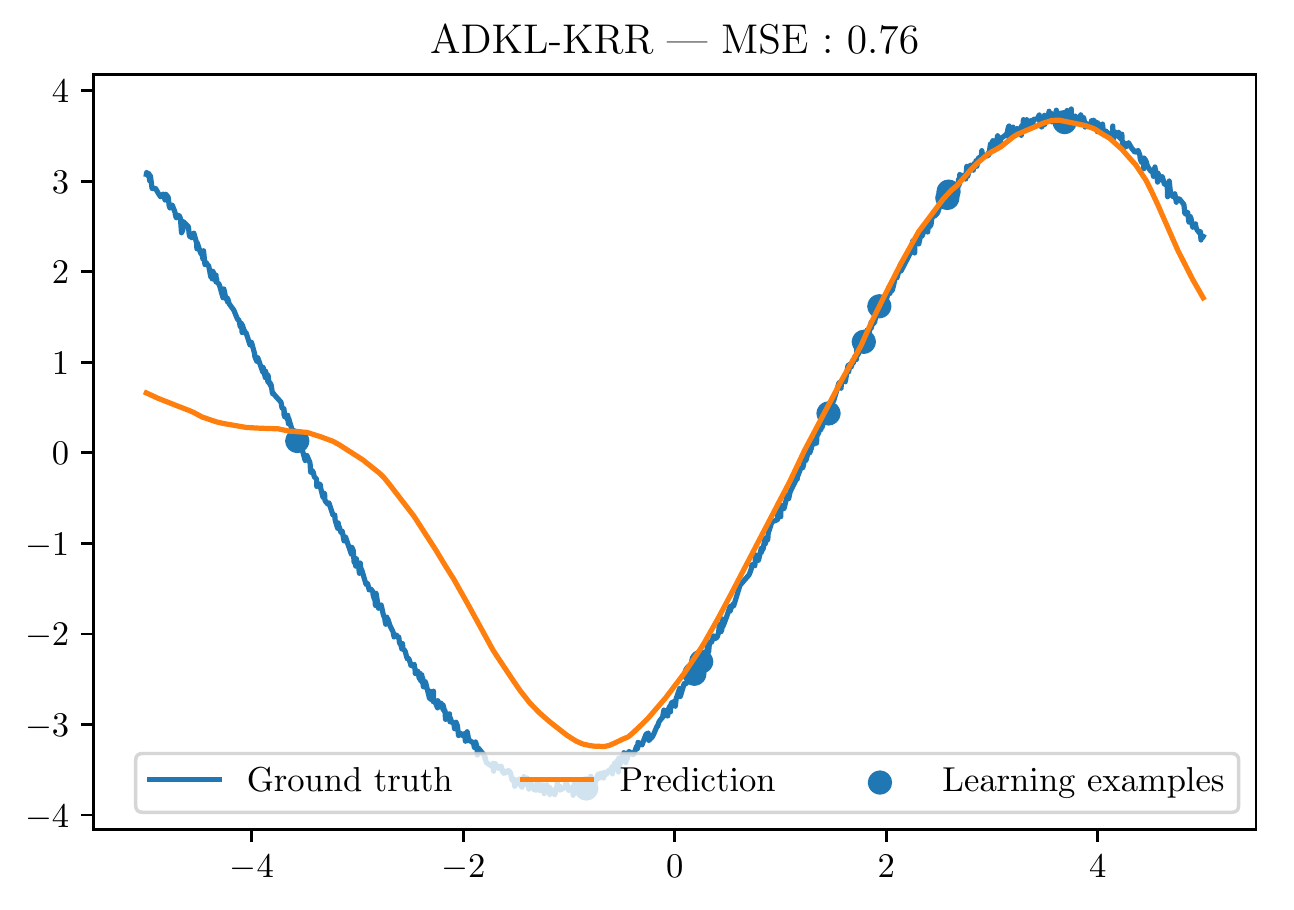}
    \end{subfigure}
    \hfill
    \begin{subfigure}{.32\textwidth}
        \centering
        \includegraphics[width=\textwidth]{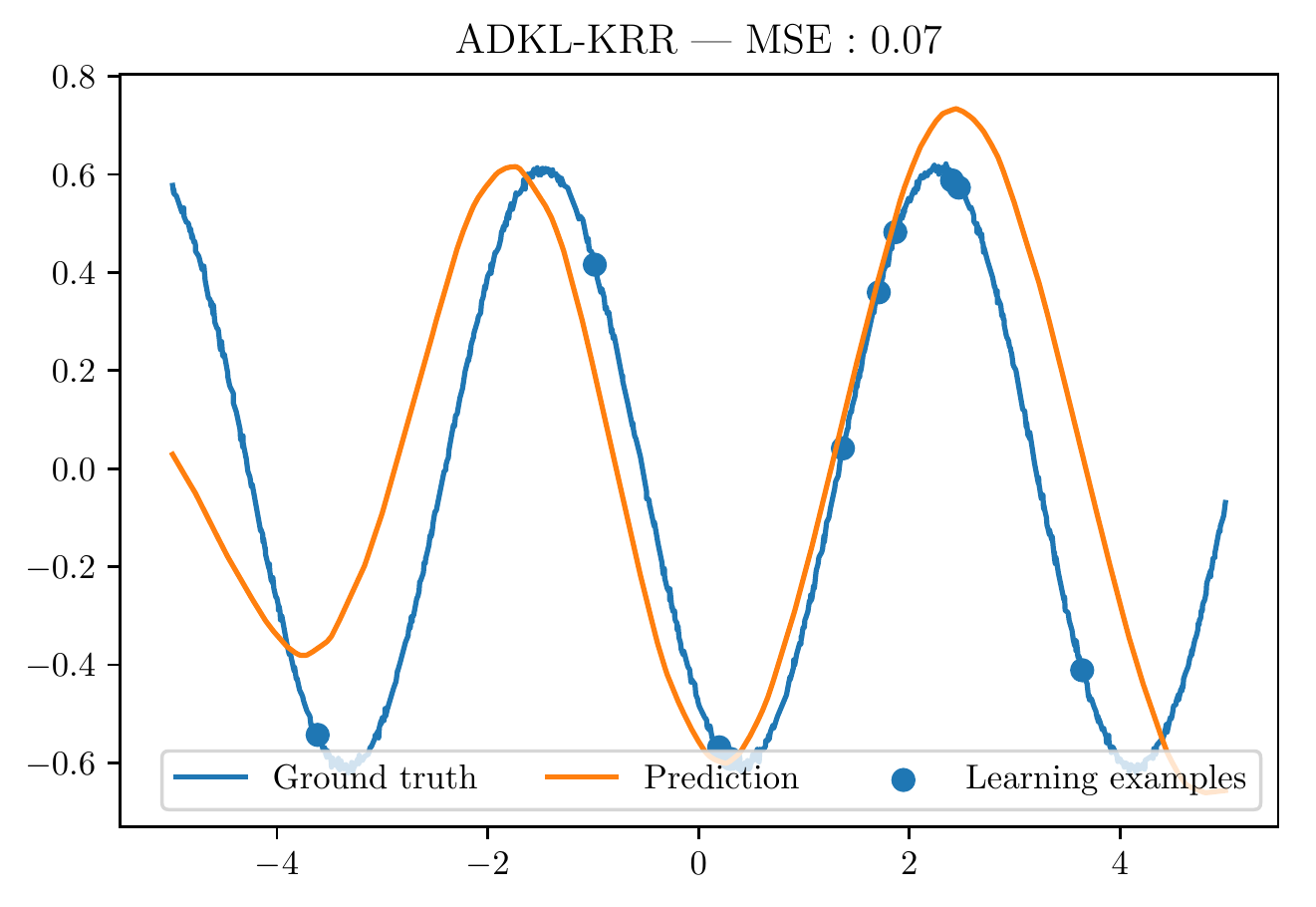}
    \end{subfigure}
    \hfill
    \begin{subfigure}{.32\textwidth}
        \centering
        \includegraphics[width=\textwidth]{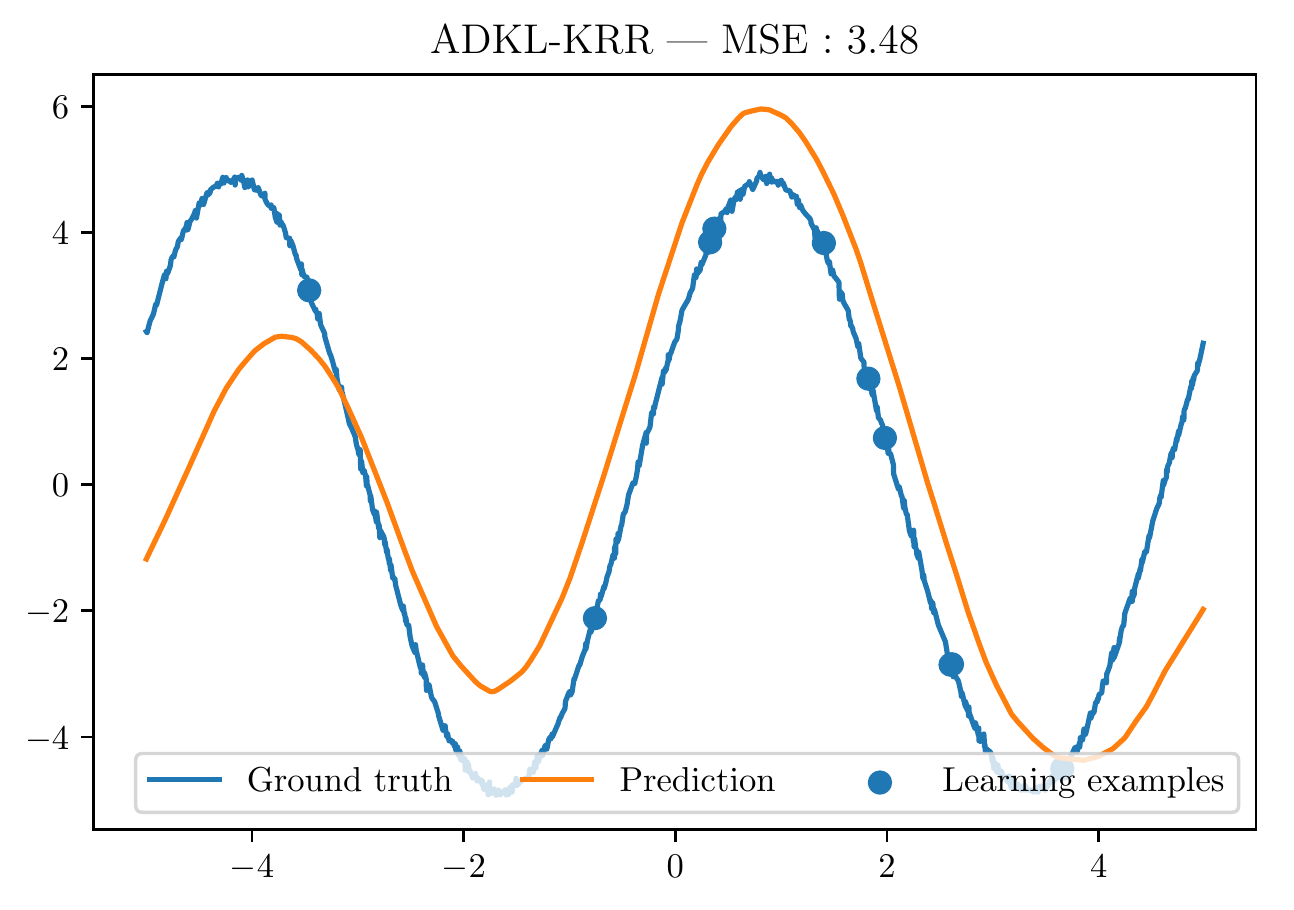}
    \end{subfigure}
    \\
    \begin{subfigure}{.32\textwidth}
        \centering
        \includegraphics[width=\textwidth]{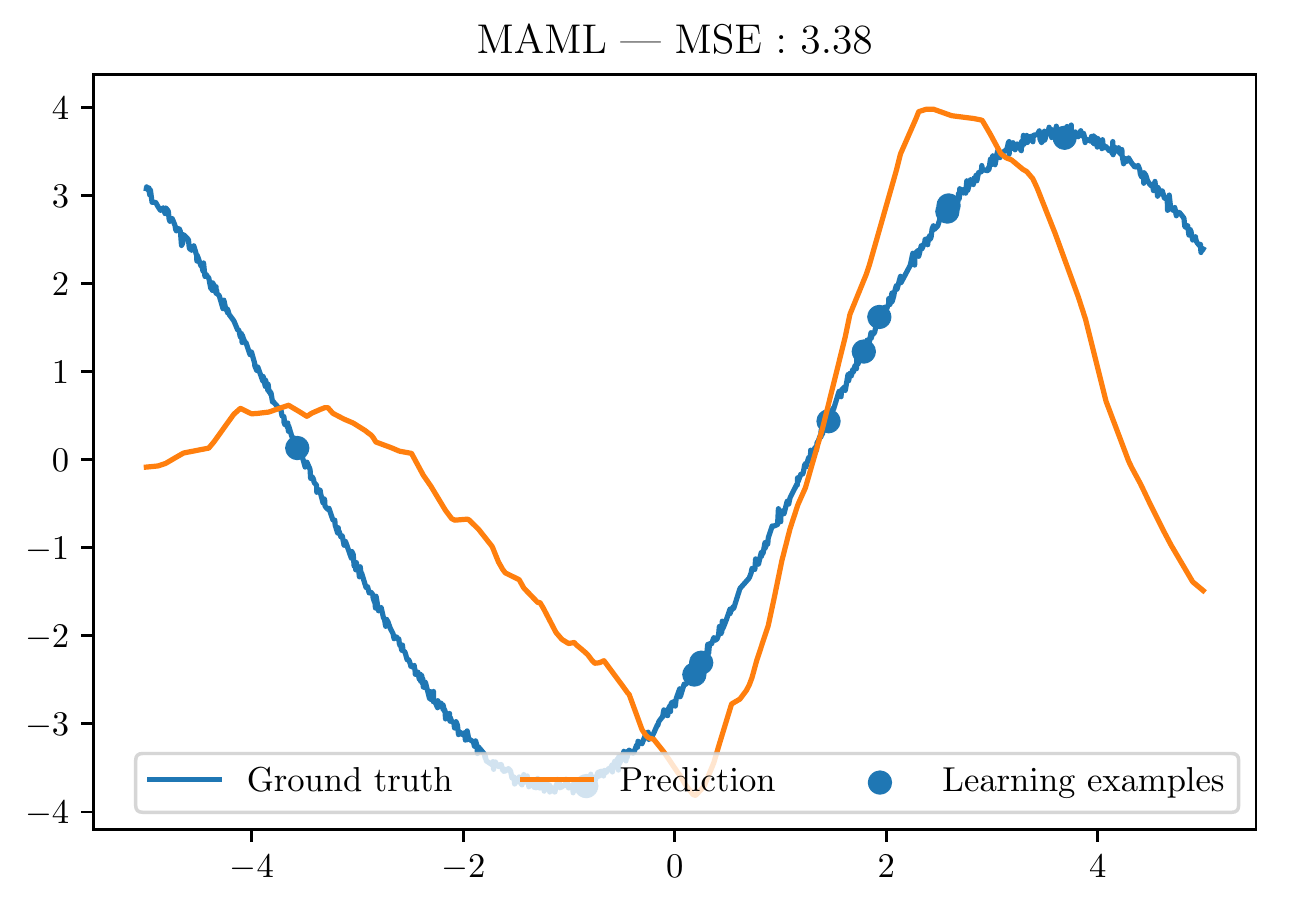}
    \end{subfigure}
    \hfill
    \begin{subfigure}{.32\textwidth}
        \centering
        \includegraphics[width=\textwidth]{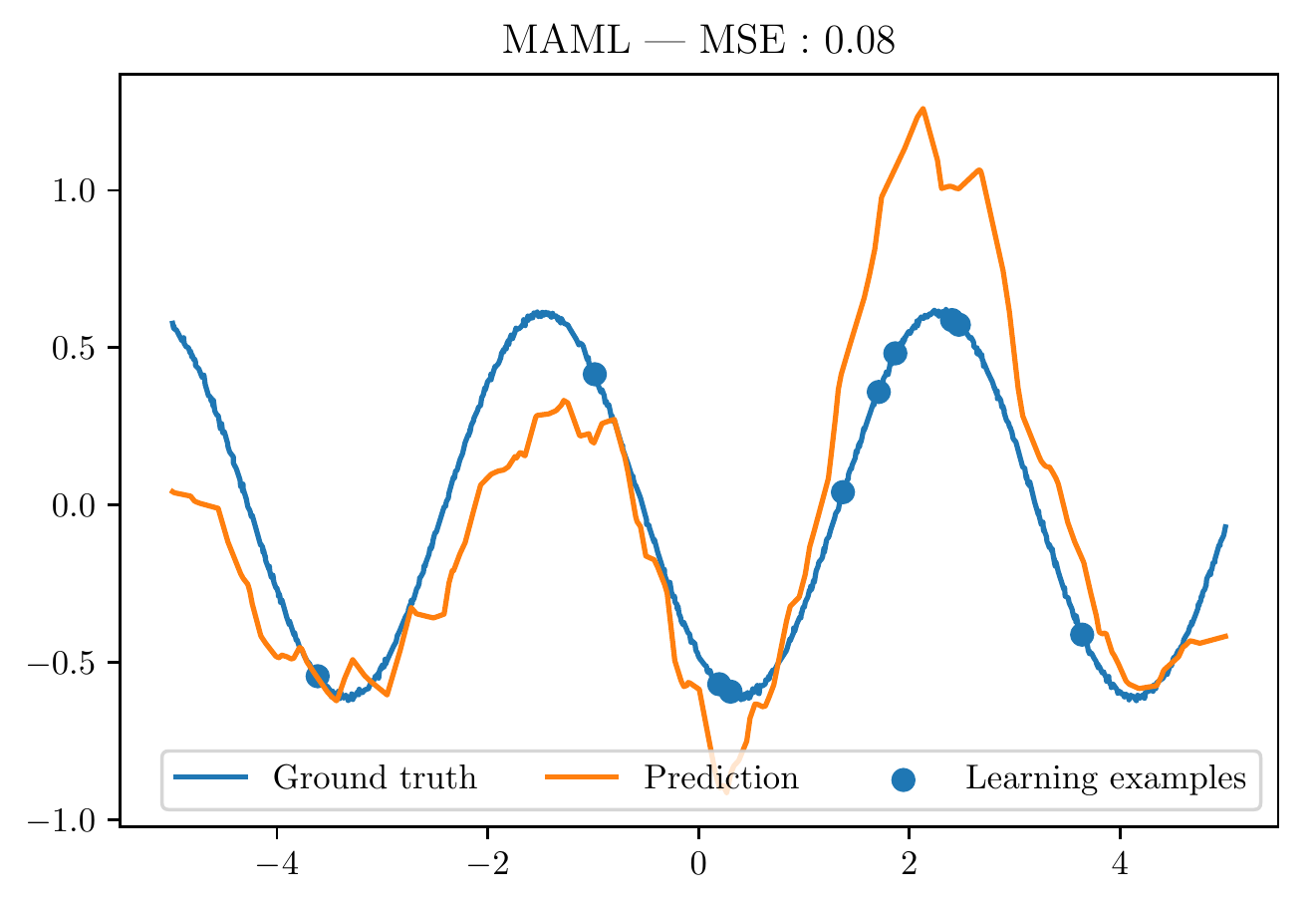}
    \end{subfigure}
    \hfill
    \begin{subfigure}{.32\textwidth}
        \centering
        \includegraphics[width=\textwidth]{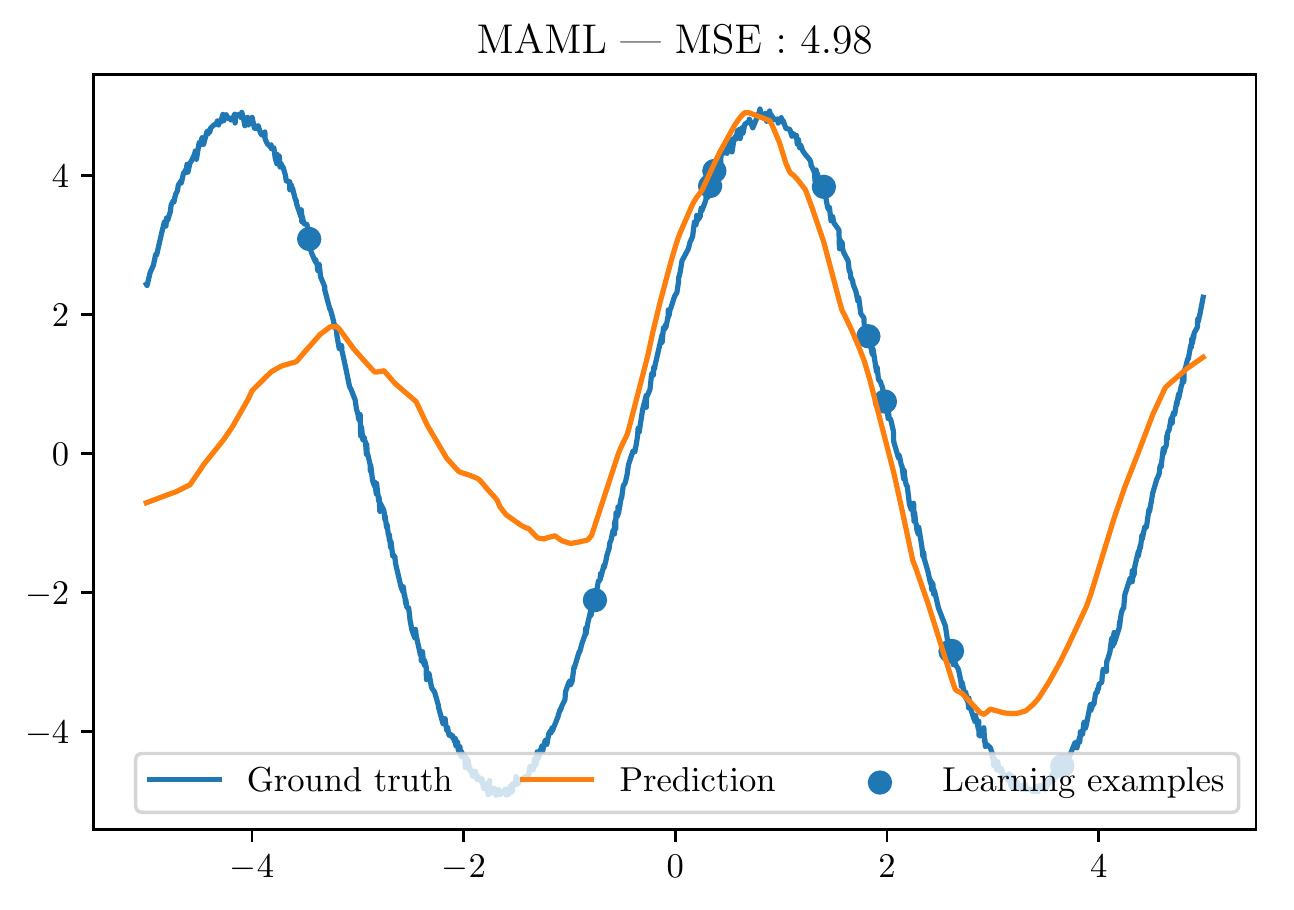}
    \end{subfigure}
    \\
    \begin{subfigure}{.32\textwidth}
        \centering
        \includegraphics[width=\textwidth]{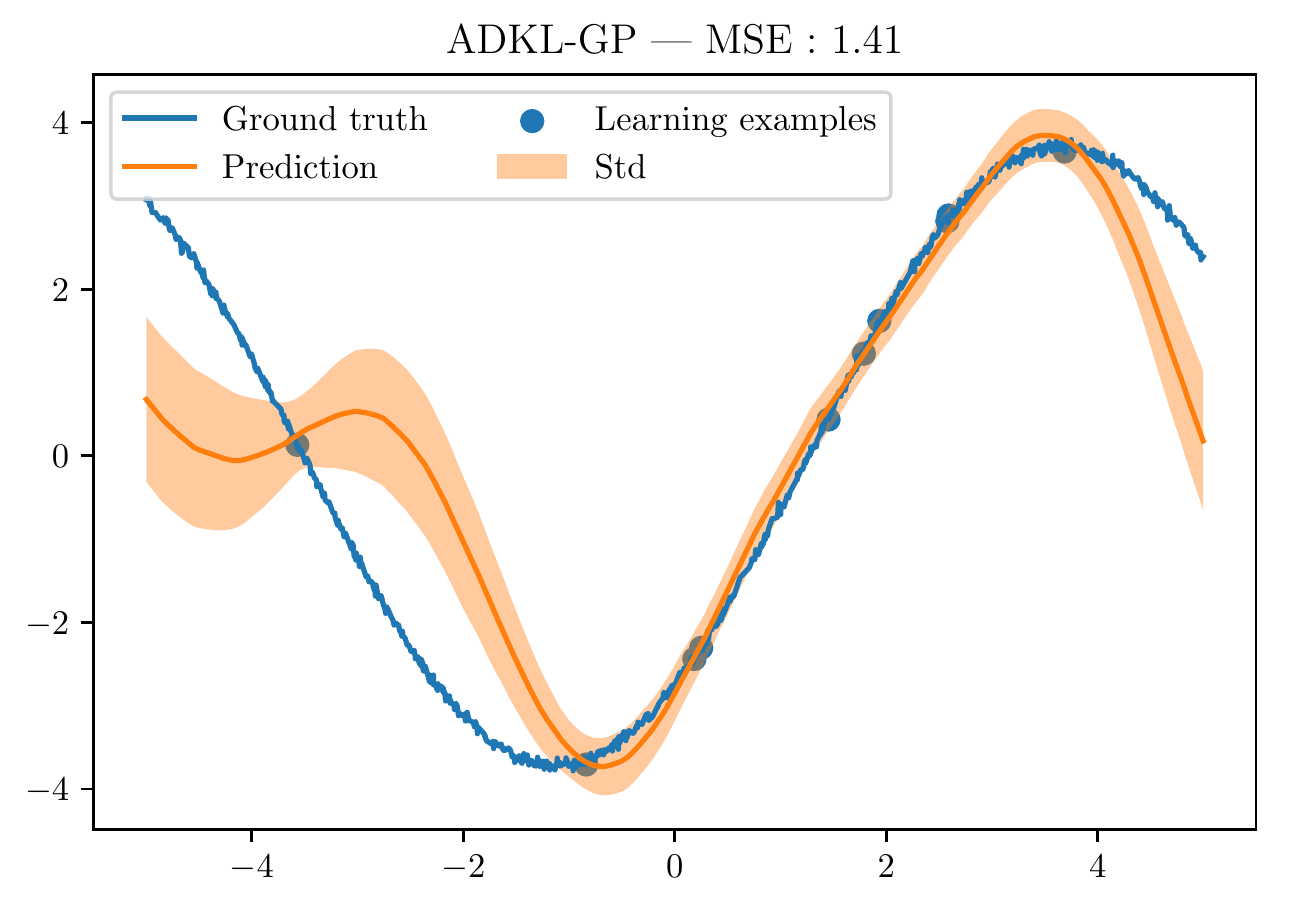}
    \end{subfigure}
    \hfill
    \begin{subfigure}{.32\textwidth}
        \centering
        \includegraphics[width=\textwidth]{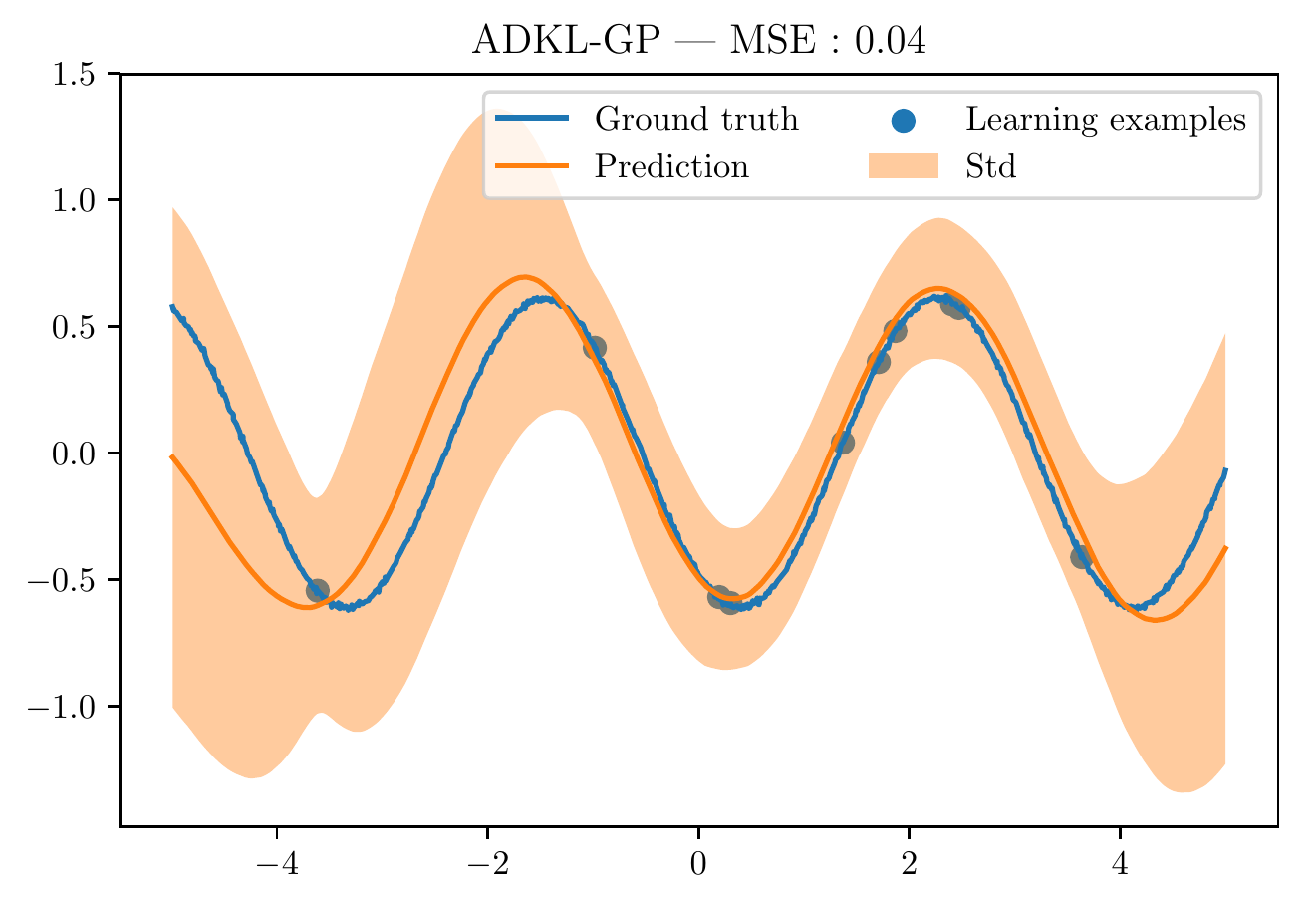}
    \end{subfigure}
    \hfill
    \begin{subfigure}{.32\textwidth}
        \centering
        \includegraphics[width=\textwidth]{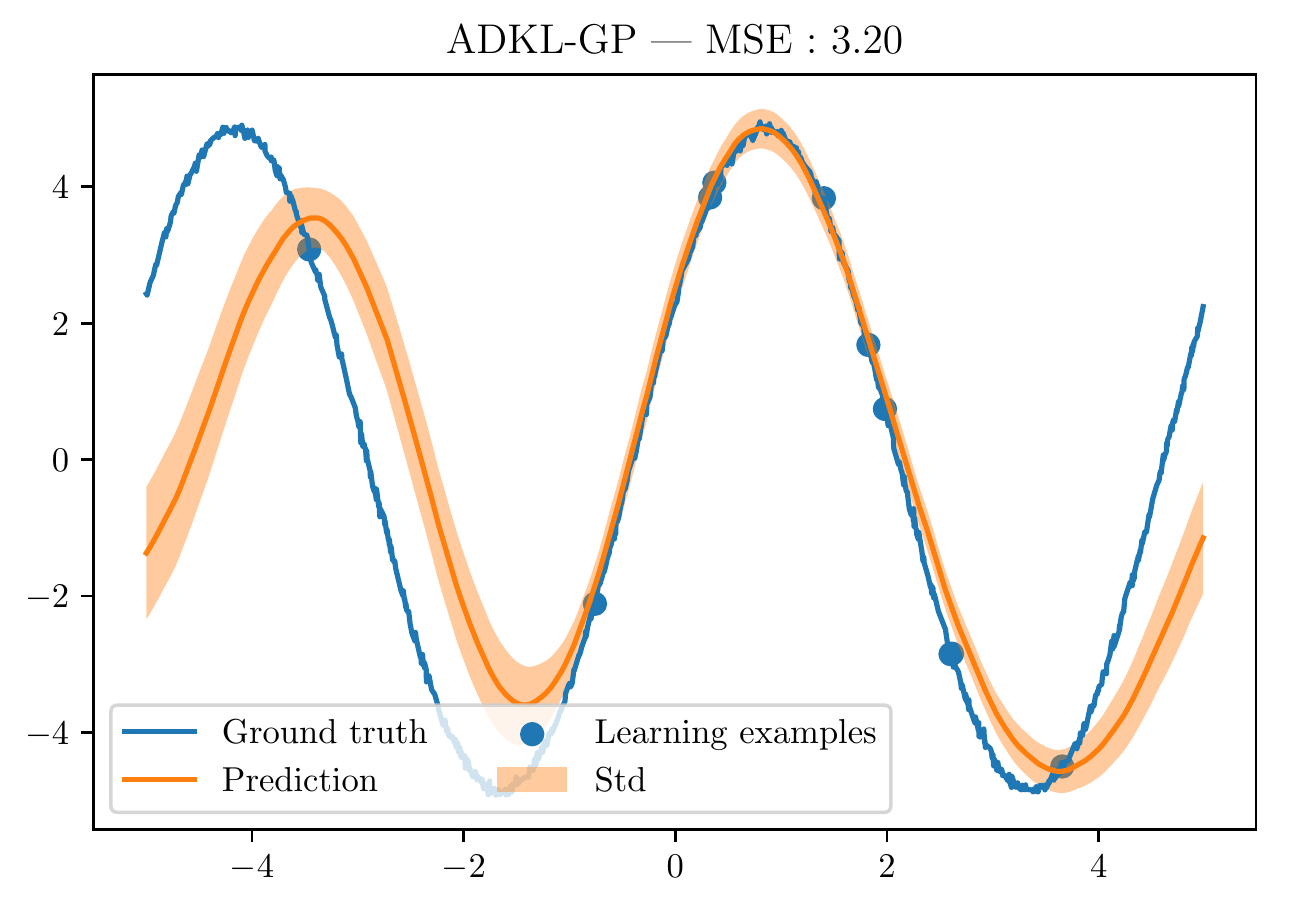}
    \end{subfigure}
    \\
    \begin{subfigure}{.32\textwidth}
        \centering
        \includegraphics[width=\textwidth]{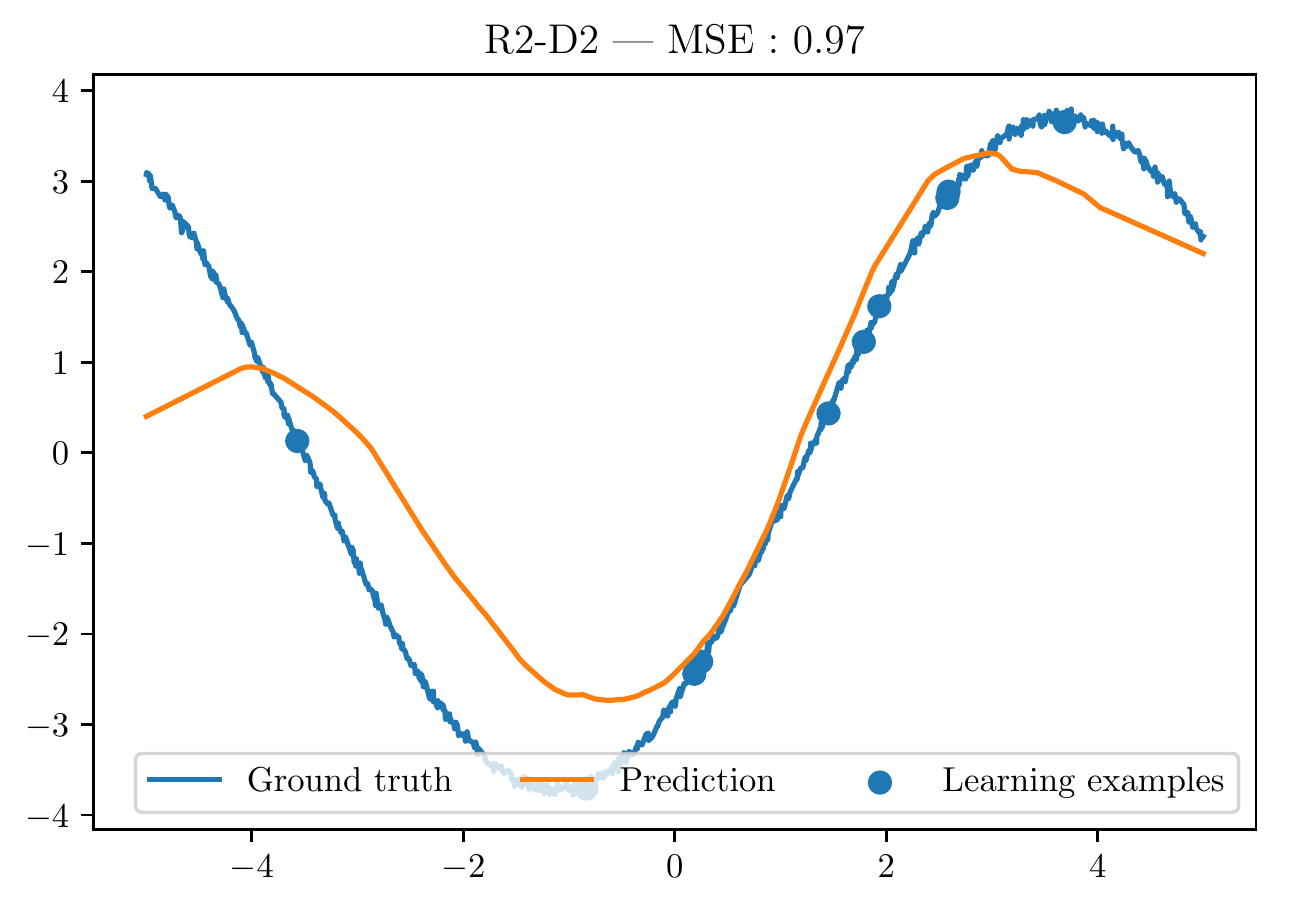}
    \end{subfigure}
    \hfill
    \begin{subfigure}{.32\textwidth}
        \centering
        \includegraphics[width=\textwidth]{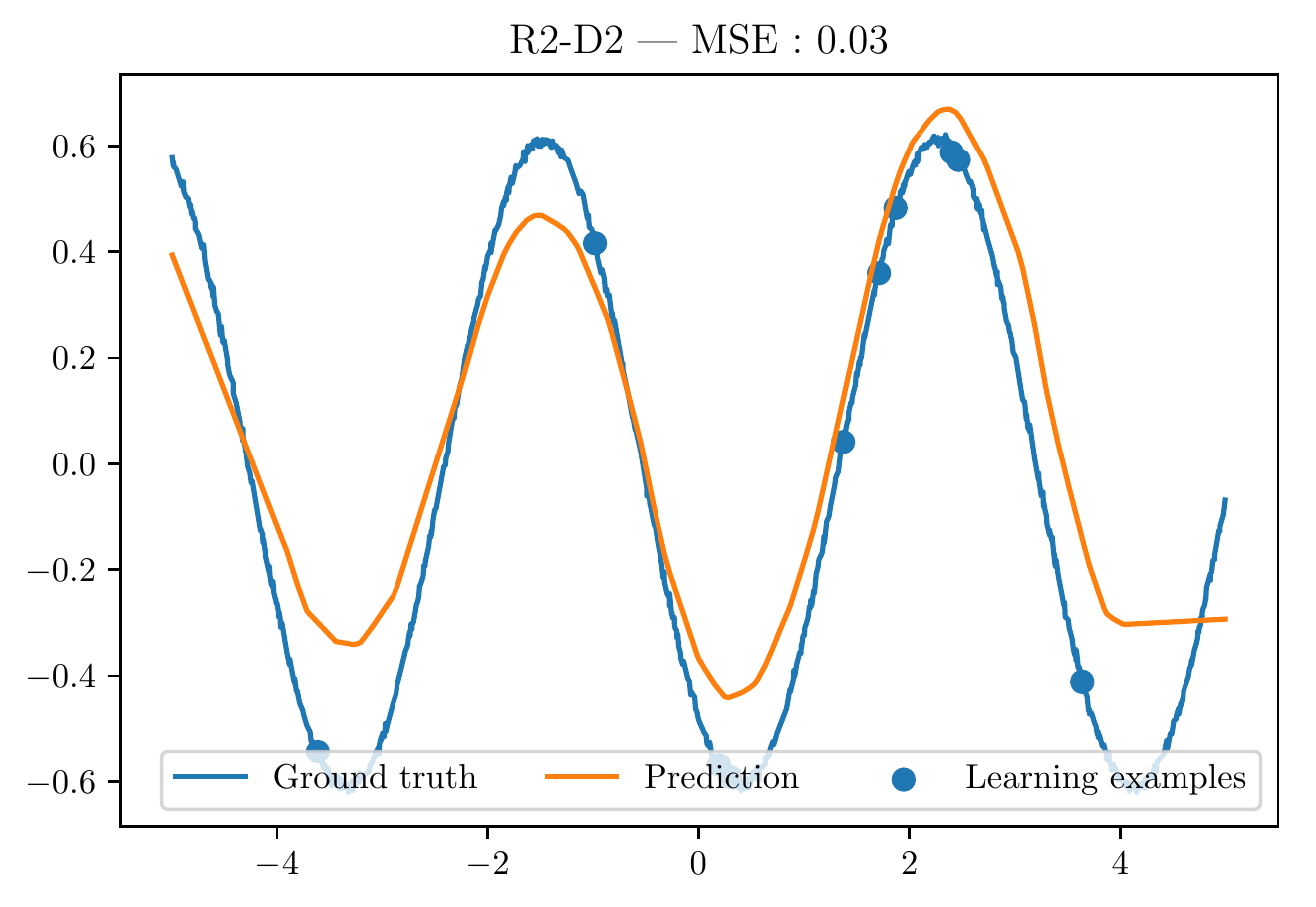}
    \end{subfigure}
    \hfill
    \begin{subfigure}{.32\textwidth}
        \centering
        \includegraphics[width=\textwidth]{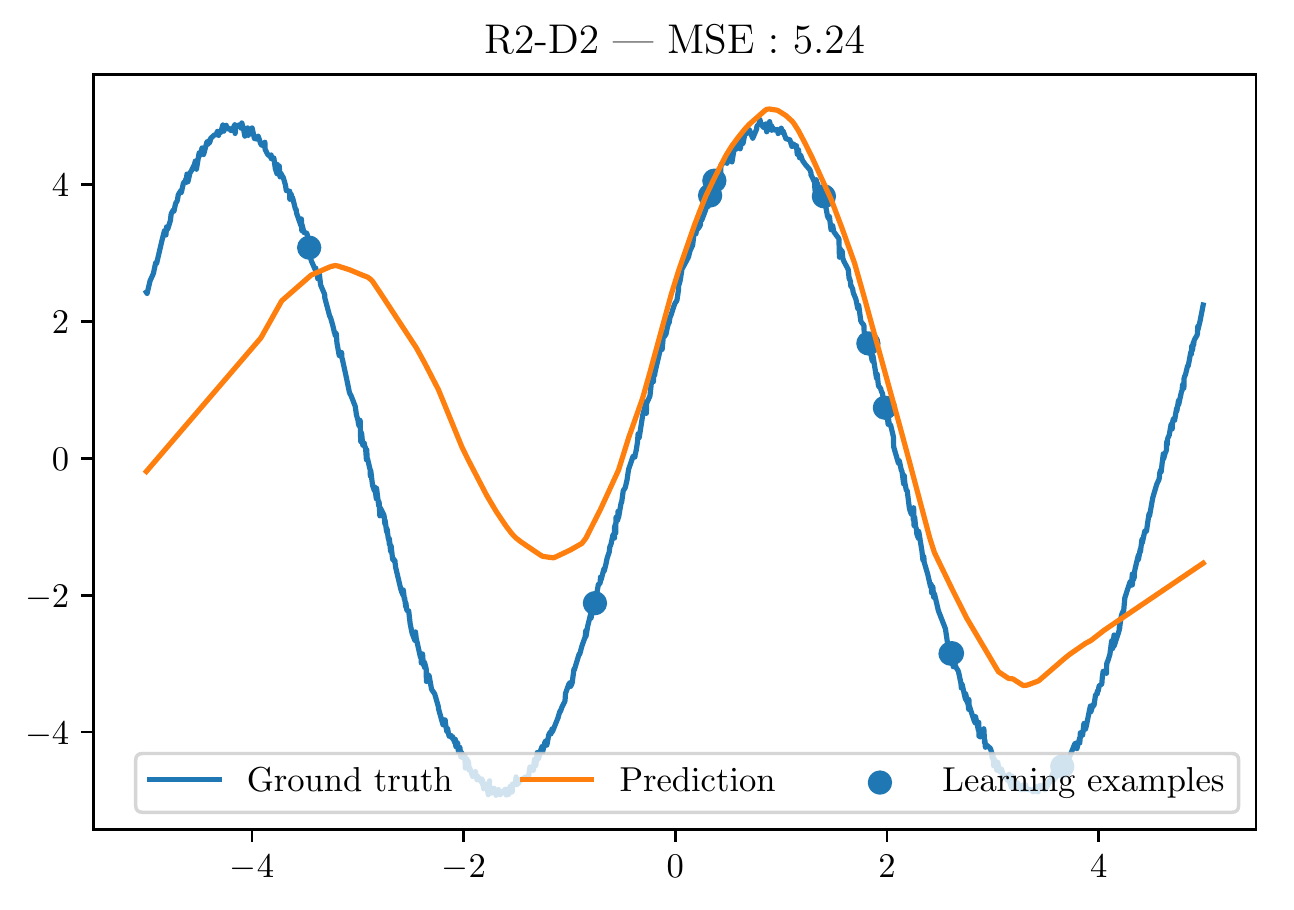}
    \end{subfigure}
    \caption{Meta-test time predictions on the \texttt{Sinusoids} collection}
    \label{fig:predictions}
\end{figure}

Figure~\ref{fig:predictions} presents a visualization of the results obtained by each model on three tasks taken from the meta-test set.
We provide the model with ten examples from an unseen task consisting of a slightly noisy sine function (shown in blue), and present in orange the the approximation made by the network based on these ten examples.

Note that contrary to others, CNP and ADKL-GP give us access to the uncertainty.

\end{document}